\documentclass[jair, 11pt]{article}
\usepackage{jair, rawfonts}

\usepackage{natbib}
\setcitestyle{authoryear}

\usepackage{lmodern}

\usepackage[utf8]{inputenc} 
\usepackage[T1]{fontenc}    
\PassOptionsToPackage{hyphens}{url}
\usepackage{url}
\usepackage{booktabs}       
\usepackage{amsfonts}       
\usepackage{nicefrac}       
\usepackage{microtype}      

\usepackage{makecell}

\usepackage{amsmath}
\usepackage{amssymb}
\usepackage{bbold}
\usepackage{enumerate}
\usepackage[dvipsnames]{xcolor}
\usepackage{pdfpages}
\usepackage{subcaption}
\usepackage{mathtools}

\usepackage{algorithm}
\usepackage{algpseudocode}

\usepackage{float}
\usepackage{bm}

\usepackage{verbatim}

\newcommand{\St}{\mathcal{S}}

\newcommand{\A}{\mathcal{A}}
\newcommand{\R}{\mathbb{R}}
\newcommand{\N}{\mathbb{N}}
\DeclareMathOperator*{\E}{\mathbb{E}}

\jairheading{73}{2022}{117-171}{09/2021}{01/2022}
\ShortHeadings{Jointly Learning Environments and Control Policies}{Bolland, Boukas, Berger, \& Ernst}
\firstpageno{117}

\title{Jointly Learning Environments and Control Policies with Projected Stochastic Gradient Ascent}

\author{\name Adrien Bolland \email adrien.bolland@uliege.be \\
 \name Ioannis Boukas \email ioannis.boukas@uliege.be \\
 \name Mathias Berger \email mathias.berger@uliege.be \\
 \name Damien Ernst \email dernst@uliege.be\\
 \addr Montefiore Institute\\
 University of Li\`ege\\
 Li\`ege, Belgium
}

\begin{document}

\maketitle

\begin{abstract}
We consider the joint design and control of discrete-time stochastic dynamical systems over a finite time horizon. We formulate the problem as a multi-step optimization problem under uncertainty seeking to identify a system design and a control policy that jointly maximize the expected sum of rewards collected over the time horizon considered. The transition function, the reward function and the policy are all parametrized, assumed known and differentiable with respect to their parameters. We then introduce a deep reinforcement learning algorithm combining policy gradient methods with model-based optimization techniques to solve this problem. In essence, our algorithm iteratively approximates the gradient of the expected return via Monte-Carlo sampling and automatic differentiation and takes projected gradient ascent steps in the space of environment and policy parameters. This algorithm is referred to as Direct Environment and Policy Search (DEPS). We assess the performance of our algorithm in three environments concerned with the design and control of a mass-spring-damper system, a small-scale off-grid power system and a drone, respectively. In addition, our algorithm is benchmarked against a state-of-the-art deep reinforcement learning algorithm used to tackle joint design and control problems. We show that DEPS performs at least as well or better in all three environments, consistently yielding solutions with higher returns in fewer iterations. Finally, solutions produced by our algorithm are also compared with solutions produced by an algorithm that does not jointly optimize environment and policy parameters, highlighting the fact that higher returns can be achieved when joint optimization is performed.
\end{abstract}

\section{Introduction} \label{sec:introduction}

Problems involving the design of a system that must be actively controlled afterwards are ubiquitous in the field of engineering. Examples range from the joint design and control of robot hands in order to achieve a specific task \citep{chen2021co} to the design and control of small-scale power systems such as microgrids that aim to provide a cheap, low-carbon source of electricity \citep{franccois2016towards}. The interplay between design and control strategies and their impact on overall system performance have long been recognized, both in artificial \citep{li2001design}, and natural systems such as the human body \citep{anderson2003embodied}. The idea that the design and control strategies of systems must be jointly optimized in order to achieve maximum performance naturally followed, and the topic has received a great deal of interest in recent years \citep{ha2017joint, dinev2021co}.

In this paper, we consider the joint design and control of discrete-time stochastic dynamical systems over a finite time horizon \citep{bertsekas2005dynamic} and formulate the problem as a multi-step optimization problem under uncertainty \citep{bakker2020structuring}. In this framework, an agent seeks to identify a system design and a control policy (i.e., a mapping between system states and control decisions) that jointly maximize the expected sum of rewards collected over the time horizon of interest (i.e., the expected return of the policy in the environment). On one hand, the transition dynamics of the system and the reward function are parametrized by so-called \textit{system} (or \textit{environment}) parameters. On the other hand, the policy is a parametric function that depends on distinct \textit{policy} parameters. Furthermore, these parametric functions are assumed to be known and to be differentiable with respect to their parameters. Solving the problem thus consists in finding the combination of system and policy parameters that jointly maximize the expected return of the policy in the environment. 

To this end, we introduce a deep reinforcement learning algorithm combining policy gradient methods with model-based optimization techniques. More specifically, in each iteration, the algorithm approximates the gradient of the expected sum of rewards with respect to both environment and policy parameters via Monte-Carlo sampling and automatic differentiation, and takes projected gradient ascent steps in the space of environment and policy parameters. The resulting algorithm is called the Direct Environment and Policy Search (DEPS) algorithm. The performance of the DEPS algorithm is empirically evaluated in three environments. The first environment is a mass-spring-damper system. The second environment deals with the design and control of an off-grid microgrid, while the third environment is concerned with the design and control of a drone. The complexity of the latter two environments is on a par with that of typical benchmark environments found in the OpenAI Gym library \citep{brockman2016openai}. In addition, for each environment, the performance of the DEPS algorithm is benchmarked against that of a state-of-the-art algorithm proposed to tackle joint design and control problems \citep{schaff2019jointly}. Finally, in order to assess the effectiveness of the joint design and control approach, we compare the outcomes of the DEPS algorithm with the outcomes produced by an algorithm that does not perform joint optimization. More precisely, this algorithm consists in designing a control policy \textit{a priori} and then optimizing environment parameters using derivative-free optimization methods so as to maximize the expected return.  

This paper is organized as follows. In Section \ref{sec:related_work}, we review the relevant literature. In Section \ref{sec:background}, we present the theoretical background and introduce the problem statement more formally. In Section \ref{sec:optimizing_system}, the proposed methodology as well as the algorithmic implementation for direct environment and policy search are described. The experimental protocol for the evaluation of the proposed algorithm is introduced and the results are discussed in Section \ref{sec:experiments}. Finally, the conclusions and future work directions are discussed in Section \ref{sec:conclusions}.

\section{Related Work} \label{sec:related_work}
In this section, we discuss the different methods that have been used for tackling the problem statement we introduced in Section \ref{sec:introduction}. We first review stochastic programming approaches and simulation-based methods relying on derivative-free optimization techniques. We then discuss reinforcement learning approaches and how policy gradient methods can be extended to tackle our problem. Finally, we briefly compare our method with existing state-of-the-art approaches from the literature.

Multi-stage stochastic programming has been widely used for a variety of applications described in the literature \citep{wallace2003stochastic}. In general, a mathematical model of the system is assumed to be available. The design and operational decisions as well as the system states are represented as optimization variables, while the system dynamics are encoded via a set of equality and inequality constraints. Some model parameters may be uncertain and may be represented as realizations of random variables or stochastic processes whose probability distributions are assumed to be known \citep{birge2011introduction}. The tractability of this approach depends on several factors, including the number of stages, the nature of the uncertainty (i.e., whether its probability distribution depends on optimization variables \citep{goel2006class} and whether the latter can be accurately approximated using a small number of scenarios \citep{Heitsch2009}), and the convexity (or lack thereof) of the resulting constrained optimization problem \citep{nemirovsky1983problem}. In practice, system design problems are often approximated via two-stage stochastic programs where the first stage represents design decisions and the second stage corresponds to operational decisions, respectively. This approach has, for instance, been applied in the context of energy systems \citep{wallace2003stochastic}, supply chains \citep{Marufuzzaman2014}, as well as robotics \citep{bravo2020one}. Once a system design has been selected, real-time operation is usually conducted using receding horizon control strategies, such as model predictive control \citep{camacho2013model}.

A common approach used in the literature to tackle joint design and control problems consists in simulating different configurations of system designs and control policies and selecting the configuration that yields the most desirable outcome \citep{brekken2010optimal}. Such methods can properly capture the impact of the uncertainty on system design and operation, provided that a sufficient number of simulations can be run. Systems with highly-nonlinear (and non-convex) dynamics are also easier to handle with such methods. Originally, these techniques relied on derivative-free optimization methods to navigate through the space of system design and policy parameters and find the combinations that perform best. However, this can be ineffective and time-consuming, especially in high-dimensional spaces \citep{jamieson2012query, oliveto2015improved}. In many problems, the number of design parameters remains moderate but the overall number of parameters grows substantially as soon as complex policies are considered (e.g., based on neural networks). As a consequence, this method is usually limited to solving problems with policies that depend on relatively few parameters. For example, in the work of \cite{digumarti2014concurrent}, the policy takes the form of an optimization problem that depends on a few hyperparameters. This optimization problem is solved at each time step in order to compute a control action that is then applied to a robot. \cite{brekken2010optimal} compare myopic rule-based policies depending on the environment parameters with a policy whose hypothesis space is defined by a small multi-layer perceptron comprising a few neurons. In both papers, derivative-free optimization methods are used to jointly update the system and policy parameters.

On the other hand, gradient descent (or ascent) methods have been very successful at learning complex function approximators with a large number of parameters (e.g., deep neural networks) for machine learning tasks such as supervised learning \citep{bottou2010large,kingma2014adam}. These ideas have also be applied in the realm of reinforcement learning (RL) \citep{kaelbling1996reinforcement}, giving rise to the class of techniques known as \textit{policy gradient} methods \citep{grondman2012survey}. The basic idea behind these methods, as formulated in the seminal work of \cite{williams1992simple} on the REINFORCE algorithm, consists in sampling trajectories from a parametrized policy, computing the gradient of the log-likelihood of the sequence of actions taken in each trajectory, and updating the policy parameters by stepping in the direction given by a weighted sum of these gradients. The weight assigned to each gradient is the cumulative reward observed in the corresponding trajectory. Hence, these methods are best viewed as model-free gradient-based optimization techniques that make it possible to learn expressive control policies for complex decision-making problems. In recent years, new policy gradient methods have met with considerable success in challenging applications dealing with high-dimensional parameter spaces such as robotics. In particular, \cite{schulman2015trust} applied the ideas of trust region methods for nonlinear optimization to reinforcement learning and derived the trust region policy optimization (TRPO) algorithm. This algorithm was then improved to yield proximal policy optimization (PPO) methods \citep{schulman2017proximal}. Roughly speaking, in PPO, the trust regions previously used in TRPO are approximated in order to reduce the computational cost of parameter updates. In the policy gradient methods discussed so far, the computation of the gradients relies on the computation of the expected return of a policy, which is usually approximated by the cumulative reward observed when playing the policy. An alternative to this approximation consists in learning the expected return using an additional function approximator. Such algorithms are referred to as \textit{actor critic} methods, where the additional function approximator is called a \textit{critic} while the policy network is referred to as the \textit{actor}. In particular, the advantage actor critic (A2C) \citep{mnih2016asynchronous} is very popular due to its effectiveness on complex problems. Another popular algorithm is the soft actor critic (SAC) algorithm \citep{haarnoja2018soft}, which takes the entropy of the policy into account for learning optimal policies. 

Building on the success of policy gradient methods, a number of researchers have proposed ways of extending them to tackle joint design and control problems and devise algorithms with a much better sample efficiency than earlier derivative-free methods. \cite{schaff2019jointly} extended the REINFORCE algorithm to account for system parameters. More precisely, their algorithm maintains a parametrized distribution over environments, which corresponds in some sense to an \textit{environment policy}, in addition to the usual control policy. Environments are then sampled from this distribution and the control policy is executed in each environment. Using information from the sampled trajectories, policy parameters are updated via PPO, while the parameters of the environment distribution are updated using the simple REINFORCE update rule. The same idea was applied in different robotic environments and was shown to perform better than other optimization techniques by \cite{ha2019reinforcement}. Another RL technique based on the SAC algorithm \citep{haarnoja2018soft} for joint design and control was proposed by \cite{luck2020data}. Here, the critic function approximator in the SAC is made explicitly dependent on the environment parameters. The update of the parametric functions (the critic and the policy) is nevertheless inefficient as it requires the maximization of the critic function with respect to environment parameters, which implies that a strongly non-convex optimization problem must be solved in each iteration.

Finally, the DEPS algorithm proposed in this paper can be interpreted as combining policy gradient methods with model-based optimization techniques to tackle joint design and control problems more effectively. It extends the REINFORCE algorithm \citep{williams1992simple} to take advantage of the knowledge of the analytical form of the environment to compute the gradients of the expected cumulative reward with respect to system parameters. The works that are closest to ours, combining policy gradients with the knowledge of system dynamics, are those of \cite{chen2020hardware} and \cite{jackson2021orchid}. First, \cite{chen2020hardware} assume that the transition dynamics of the decision process depend on environment parameters. Then, they express the dependence of the cumulative reward on these parameters via a computational graph and use this graph to learn a joint distribution over environment parameters and control actions using a variant of the REINFORCE algorithm \citep{williams1992simple}. In the same vein, \cite{jackson2021orchid} directly learn the parameters of the dynamics jointly with those of a control policy. This is achieved by performing gradient descent on a variant of the loss function defined in the A2C algorithm, which is made explicitly dependent on the environment parameters. The method of \cite{jackson2021orchid} is more flexible than that of \cite{chen2020hardware}, as it does not require to explicitly encode the dependence of the cumulative reward on system parameters in a computational graph. In addition, it directly learns system parameters instead of learning a distribution over these parameters. Nevertheless, both methods differ from the DEPS algorithm as they only assume that the system dynamics are parametrized. The latter assumption is restrictive for modeling a large class of problems where the decision concerning the design is associated with an explicit reward or cost (e.g., to represent some notion of investment cost).

\section{Theoretical Background and Problem Statement} \label{sec:background}

In this section, we provide a generic formulation for the problem of controlling a discrete-time stochastic dynamical system over a finite time horizon. We then introduce parametrized environments and parametrized policies. Subsequently, we formulate the problem of jointly optimizing the environment and policy parameters so as to maximize the expected sum of rewards.

\subsection{Discrete-Time Dynamical Systems} \label{sec:system_def}

Let us consider the problem of controlling a discrete-time (time-invariant) stochastic dynamical system over a finite time horizon \citep{bertsekas2005dynamic}. Our formulation relies on two crucial elements, namely a state-space model representing the system dynamics and a reward function defining the control objective. Let $T\in \N$ be the optimization horizon, which refers to the number of decision stages in the control process. The system is defined by a state space $\St$, an action space $\A$, a disturbance space $\Xi$, a transition function $f : \St \times \A \times \Xi \rightarrow \St$, and a conditional probability distribution $P_\xi$ giving the probability $P(\xi_t|s_t, a_t)$ of drawing a disturbance $\xi_t \in \Xi$ when taking an action $a_t \in \A$ while being in a state $s_t \in \mathcal{S}$. A probability measure $P_0$ yields the probability $P_0(s_0)$ of each state $s_0 \in \St$ to be the initial state. At time $t \in \{0, 1, \dots, T-1 \}$, the system moves from state $s_t \in \St$ to state $s_{t+1} \in \St$ under the effect of an action $a_t \in \A$ and a random disturbance $\xi_t \in \Xi$, which is captured by the transition function $f$: 
\begin{align}
s_{t+1} = f(s_t, a_t, \xi_t)  \; .
\end{align}
Moreover, a reward function $\rho : \St \times \A \times \Xi \rightarrow R \subsetneq\R$ modelling the control objective is also defined and a reward $r_t = \rho(s_t, a_t, \xi_t)$ is collected after each state transition. We consider bounded reward functions, such that $|r_t| \leq r_{max}$. The different elements of this optimal control problem are gathered in a tuple $(\St, \A, \Xi, P_0, f, \rho, P_\xi, T)$, which is referred to as the environment.

We define a closed-loop policy $\pi \in \Pi$ as a function associating a probability distribution with support $\A$ to the state $s_t$ of the system at decision stage $t=0, \dots, T-1$. At each iteration, an action $a_t$ is sampled from the policy with probability $\pi(a_t | s_t, t)$ and applied to the dynamical system, giving rise to a state transition. A trajectory $\tau = (s_0, a_0,\xi_0, a_1, \xi_1, \dots a_{T-1}, \xi_{T-1})$ contains the information collected by executing policy $\pi$ over the horizon $T$. The cumulative reward $R(\tau)$ over trajectory $\tau$ can be computed as:
\begin{align}
R(\tau) &= \sum_{t=0}^{T-1} \rho(s_t, a_t, \xi_t)  \; , \label{eq:cum_rew_traj}
\end{align}
where $s_{t+1} = f(s_t, a_t, \xi_t)$.
The expected cumulative reward associated with a policy $\pi$ and a state $s_t \in\St$ at time $t$ is called the return of the policy and is given by: 
\begin{align}
V^{\pi}(s_t, t) &= \sum_{t'=t}^{T-1} \E_{\underset{\xi_{t'} \sim P_\xi(\cdot |s_{t'}, a_{t'})}{a_{t'} \sim \pi(\cdot | s_{t'}, t')}} \{\rho(s_{t'}, a_{t'}, \xi_{t'})\}  \; . \label{eq:def_v}
\end{align}
Optimal policies are defined by the \textit{principle of optimality} \citep{bertsekas2005dynamic}. This principle states that a policy is optimal in state $s_t$ at time $t$ if it maximizes the expected reward-to-go from that state at that time. An optimal policy $\pi^* \in \Pi$ is thus such that, $\forall s_t \in \St, \forall t \in \{0, \dots, T-1\}$:
\begin{align}
	\pi^* &\in  \underset{\pi \in \Pi}{\mathrm{argmax}} \{V^{\pi}(s_t, t) \} \; .
\end{align}

\subsection{Problem Statement: Optimizing over a Set of Environments and Policies} \label{sec:problem_statement}
We consider a parametrized version of the environment defined in Section \ref{sec:system_def}, with continuous state space $\St \subsetneq\R^{d_\St}$, $d_\St \in \mathbb{N}$, action space $\A \subsetneq\R^{d_\A}$, $d_\A \in \mathbb{N}$, disturbance space $\Xi \subsetneq\R^{d_\Xi}$, $d_\Xi \in \mathbb{N}$, distribution $P_0$ over the initial states and horizon $T$. The state, action and disturbance spaces are assumed to be compact. Moreover, the transition and reward functions $f_\psi$ and $\rho_\psi$ are parametrized by the vector $\psi$ defined over the compact $\Psi \subsetneq\R^{d_\Psi}$, $d_\Psi \in \mathbb{N}$. Both functions are assumed continuously differentiable on the parameter space $\Psi$ and the state space $\St$ for every action in $\A$ and every disturbance in $\Xi$. Furthermore, the disturbance distribution $P_\xi$ is assumed to be differentiable on the state space $\St$ for every action in $\A$  and every disturbance in $\Xi$. Additionally, the policy $\pi_\theta$ is parametrized by a real vector $\theta$ taking values in the compact $\Theta \subsetneq\R^{d_\Theta}$, $d_\Theta \in \mathbb{N}$. The policy is assumed continuously differentiable on its parameter space $\Theta$ and on the state space $\St$ for every action in $\A$ and every time $t$. We seek to identify a pair of parameter vectors $(\psi, \theta)$ such that the policy $\pi_\theta$ maximizes the expected return, on expectation over the initial states, in the environment $(\St, \A, \Xi, P_0, f_\psi, \rho_\psi, P_\xi, T)$. We thus aim to solve the following optimization problem:
\begin{align}
\psi^*, \theta^* &\in \underset{\psi \in \Psi, \theta \in \Theta}{\mathrm{argmax\;}} V(\psi, \theta) \label{eq:problem_statement} \\
V(\psi, \theta)
&= \E_{\underset{\xi_t \sim P_\xi(\cdot|s_t, a_t)}{\underset{a_t \sim \pi_\theta(\cdot|s_t, t)}{s_0\sim P_0(\cdot)}}} \{ \sum_{t=0}^{T-1} r_t \} \label{eq:exp_cum_rew_param} \\
s_{t+1} &= f_\psi(s_t, a_t, \xi_t) \\
r_t &= \rho_\psi(s_t, a_t, \xi_t) \; .
\end{align}

\section{Direct Environment and Policy Search} \label{sec:optimizing_system}

In this section, we address the problem defined in Section \ref{sec:problem_statement}. First, we show in Section \ref{sec:gradient_computation} that the expected return is differentiable with respect to the parameters of the environment and the policy if the different parametric functions and the disturbance probability function are continuously differentiable. In such a context, we derive an analytical expression for the gradient of the expected return. The results are also extended to discrete action and disturbance spaces. In addition, we derive the expression of an unbiased estimator of the gradient from the differentiation of a loss function built from Monte-Carlo simulations. In Section \ref{sec:projected_gradien_ascent}, we discuss update rules for environment and policy parameters using the aforementioned gradient estimator and projection operators. Finally, our Direct Environment and Policy Search (DEPS) algorithm is detailed in section \ref{sec:DEPS}.

\subsection{Environment and Policy Gradients} \label{sec:gradient_computation}
In this section, we derive an analytical expression for the gradients of the expected cumulative reward of a policy with respect to the parameters of the environment and the policy. Furthermore, we explain how we can use automatic differentiation (also referred to as autodifferentiation or autodiff) to compute an unbiased estimator of these gradients in practice.

In Theorem 1, we first prove the differentiability of the expected cumulative reward with respect to the policy and the environment parameters, under the assumption that the functions used to build the environment and the policy are continuously differentiable. We then extend these results in a straightforward way to the case where $\mathcal{A}$ and/or $\Xi$ are discrete in Corollary 1. Corollaries 2 and 3 finally give the expressions of the gradients.

\paragraph{Theorem 1.} Let $(\St, \A, \Xi, P_0, f_\psi, \rho_\psi, P_\xi, T)$ and $\pi_\theta$ be an environment and a policy as defined in Section \ref{sec:problem_statement}. Additionally, let the functions $f_\psi$, $\rho_\psi$ and $P_\xi$ be continuously differentiable on the parameter space $\Psi$ and on the state space $\St$ for every action in $\A$ and every disturbance in $\Xi$. Furthermore, let the policy $\pi_\theta$ be continuously differentiable on its parameter space $\Theta$ and on the state space $\St$ for every action in $\A$ and for every time $t$. Let $V(\psi, \theta)$ be the expected cumulative reward of policy $\pi_\theta$, averaged over the initial states, for all $(\psi, \theta) \in \Psi\times\Theta$, as defined in equation \eqref{eq:exp_cum_rew_param}.
Then, the function $V$ exists, is bounded, and is continuously differentiable in the interior of $\Psi\times\Theta$.
 
\paragraph{Corollary 1.} The function $V$, as defined in Theorem 1, exists, is bounded, and is continuously differentiable in the interior of $\Psi\times\Theta$ if $\A$ and/or $\Xi$ are discrete.
 
\paragraph{Corollary 2.} The gradient of the function $V$ defined in equation \eqref{eq:exp_cum_rew_param} with respect to the parameter vector $\psi$ is such that:
\begin{multline}
\nabla_\psi  V(\psi, \theta)
= \E_{\underset{\xi_t \sim P_\xi(\cdot|s_t, a_t)}{\underset{a_t \sim \pi_\theta(\cdot|s_, t)}{s_0\sim P_0(\cdot)}}} \Big \{\Big (\sum_{t=0}^{T-1} \big( \nabla_s \log \pi_\theta(a_t|s, t)|_{s=s_t} + \nabla_s \log P_\xi(\xi_t| s, a_t)|_{s=s_t} \big ) \cdot \nabla_\psi s_t  \Big ) \\
\times \Big (\sum_{t=0}^{T-1} r_t \Big ) + \Big (\sum_{t=0}^{T-1} \nabla_\psi \rho_\psi(s, a_t, \xi_t)|_{s=s_t} + \nabla_s \rho_\psi(s, a_t, \xi_t)|_{s=s_t} \cdot \nabla_\psi s_t \Big ) \Big \} \label{eq:grad_v_psi}  ,
\end{multline}
where:
\begin{align}
\nabla_\psi s_t &= (\nabla_s f_\psi)(s, a_{t-1}, \xi_{t-1})|_{s=s_{t-1}} \cdot \nabla_\psi s_{t-1} + (\nabla_\psi f_\psi)(s, a_{t-1}, \xi_{t-1})|_{s=s_{t-1}} \label{eq:grad_s}  \; ,
\end{align}
with $\nabla_\psi s_0 = 0$.

\paragraph{Corollary 3.} The gradient of the function $V$, defined in equation \eqref{eq:exp_cum_rew_param}, with respect to the parameter vector $\theta$ is given by:
\begin{align}
\nabla_\theta  V(\psi, \theta)
&=  \E_{\underset{\xi_t \sim P_\xi(\cdot|s_t, a_t)}{\underset{a_t \sim \pi_\theta(\cdot|s_t, t)}{s_0\sim P_0(\cdot)}}} \{ (\sum_{t=0}^{T-1} \nabla_\theta \log \pi_\theta(a_t|s_t, t)) (\sum_{t=0}^{T-1} r_t) \} \label{eq:grad_v_theta} \;  .
\end{align}

\paragraph{Definition 1.} Let $(\St, \A, \Xi, P_0, f, \rho, P_\xi, T)$ and $\pi$ be an environment and a policy, respectively, as defined in Section \ref{sec:background}. A history $h$ of the policy in the environment is a sequence:
\begin{align}
h = (s_0, a_0, \xi_0, r_0, a_1, \xi_1, r_1, \dots a_{T-1}, \xi_{T-1}, r_{T-1}) \; ,
\end{align}
where $s_0$ is an initial state sampled from $P_0$, and where, at time $t$, $\xi_t$ is a disturbance sampled from $P_\xi$, $a_t$ is an action sampled from $\pi$,  and $r_t$ is the reward observed. 

The DEPS algorithm will exploit the following theorem that shows that an unbiased estimate of the gradients of the expected cumulative reward can be obtained by evaluating the gradients of a loss function computed from a set of histories. Automatic differentiation will later be used for computing these gradients in our simulations.  

\paragraph{Theorem 2.} Let $(\St, \A, \Xi, P_0, f_\psi, \rho_\psi, P_\xi, T)$ and $\pi_\theta$ be an environment and a policy, respectively, as defined in Section \ref{sec:problem_statement}. Let $V(\psi, \theta)$ be the expected cumulative reward of policy $\pi_\theta$ averaged over the initial states, as defined in equation \eqref{eq:exp_cum_rew_param}. Let $\mathcal{D} = \{h^m| m = 0, \dots, M-1\}$ be a set of $M$ histories sampled independently and identically from the policy $\pi_\theta$ in the environment. Let $\mathcal{L}$ be a loss function such that,  $\forall (\psi, \theta) \in \Psi\times\Theta$:
\begin{multline}
\mathcal{L}(\psi, \theta) = -\frac{1}{M} \sum_{m=0}^{M-1} \Big (\sum_{t=0}^{T-1} \log \pi_\theta(a_t^m|s_t^m, t) + \log P_\xi(\xi_t^m| s_t^m, a_t^m) \big ) 
\\ \times \big ((\sum_{t=0}^{T-1} r_t^m) - B \big ) 
+ \big (\sum_{t=0}^{T-1} \rho_\psi(s_t^m, a_t^m, \xi_t^m) \big) \Big)   \label{eq:loss_function} \; ,
\end{multline}
where $B$ is a constant value called the baseline.
The gradients with respect to $\psi$ and $\theta$ of the loss function are unbiased estimators of the gradients of the function $V$ as defined in equation \eqref{eq:exp_cum_rew_param} with opposite directions, i.e., they are such that: 
\begin{align}
\E_{\underset{\xi_t \sim P_\xi(\cdot|s_t, a_t)}{\underset{a_t \sim \pi_\theta(\cdot|s_t, t)}{s_0\sim P_0(\cdot)}}} \{ \nabla_\psi \mathcal{L}(\psi, \theta)\} &= - \nabla_\psi  V(\psi, \theta) \\
\E_{\underset{\xi_t \sim P_\xi(\cdot|s_t, a_t)}{\underset{a_t \sim \pi_\theta(\cdot|s_t, t)}{s_0\sim P_0(\cdot)}}} \{ \nabla_\theta \mathcal{L}(\psi, \theta) \} &= - \nabla_\theta  V(\psi, \theta)  \; .
\end{align}

\paragraph{Corollary 4.} The gradient of the loss function, defined in equation \eqref{eq:loss_function}, with respect to $\theta$ corresponds to the opposite of the update direction computed with the REINFORCE algorithm \citep{williams1992simple} averaged over $M$ simulations.

The proofs for the theorems and corollaries presented in this section are given in Appendix \ref{ap:proof}.

\subsection{Projected Gradient Ascent}\label{sec:projected_gradien_ascent}
In this section, we discuss gradient-based update rules for the environment and policy parameters. These rules make use of projection operators \citep{cohen2016projected} in order to account for constraints on the values that parameters can take.

Gradient ascent is an iterative optimization technique where optimization variables are updated in each iteration $k$ by taking a step of pre-specified size $\alpha$ in the direction provided by the gradient of the objective function with respect to optimization variables. In machine learning applications, the step size is also called the \textit{learning rate}. In the problem defined in equation \eqref{eq:problem_statement}, we aim to find a parameter vector $x = (\psi, \theta) \in X = \Psi \times \Theta \subsetneq\mathbb{R}^{d_{\Psi}+ d_{\Theta}}$ that maximizes the expected return. Standard gradient ascent thus consists in updating the parameter vector $x_k$ in iteration $k$ using the following rule:
\begin{align}
x_{k+1} &\leftarrow x_{k} + \alpha \cdot \nabla_x V(x_{k})  \; . \label{eq:grad_update}
\end{align} 

However, $x_{k+1}$ may not always belong to the constraint set $X$. In projected gradient ascent, we choose the point in $X$ that is the closest to $x_{k+1}$ according to the Euclidean distance. The projection $\Pi_{X}(y)$ of a point $y$ onto a set $X$ is defined as:
\begin{align}
\Pi_{X}(y) = \arg \min_{x\in X} \frac{1}{2}\parallel x - y \parallel_{2}^{2} \; .
\end{align}
Using projected gradient ascent, we first compute the update: 
\begin{align}
y_{k+1} &= x_{k} + \alpha \cdot \nabla_x V(x_{k}) \; , \label{eq:grad_update_step}
\end{align} 
and we then project $y_{k+1}$ onto the feasible set $X$:
\begin{align}
x_{k+1} &\in \Pi_{X} (y_{k+1} ) \; .\label{eq:proj_step}
\end{align} 
The computational cost of a projected gradient descent (or ascent) step largely depends on the amount of effort required to compute the projection in equation \eqref{eq:proj_step}, which itself depends on the properties of $X$. In practice, this approach works best with simple constraint sets for which a closed-form of the projection is readily available and inexpensive (e.g., hypercubes, for which a projection boils down to clipping the values of variables). 


\subsection{Optimizing Environment and Policy Parameters} \label{sec:DEPS}
In this section, we detail the DEPS algorithm that combines the computation of the gradients of the expected return with projected gradient ascent to find the parameters of an environment and a policy as described in the problem statement.

The DEPS algorithm iteratively updates the vectors of parameters $\psi$ and $\theta$ using projected gradient ascent according to equations \eqref{eq:grad_update_step} and \eqref{eq:proj_step}. The gradients are approximated using Theorem 2. In practice, we thus generate histories of the policy in the environment to compute the loss function \eqref{eq:loss_function}. Furthermore, the baseline is taken as the estimate of the expected cumulative reward obtained by averaging the observed cumulative reward over the $M$ histories $h^m$ used for computing the loss function:
\begin{align}
	B &= \frac{1}{M} \sum_{m=0}^{M-1} \sum_{t=0}^{T-1} r_t^m \; .
\end{align}
A Monte-Carlo approximation of the exact gradients of the loss function with respect to $\psi$ and $\theta$ is obtained via automatic differentiation. This procedure is referred to as projected \textit{stochastic} gradient ascent, since a Monte-Carlo approximation of the gradients is used. Let us remark that, in practice, we assume that the gradients exist on the boundary of $\Psi \times \Theta$. If this assumption does not hold, we can consider a compact subset $K$ of the interior of $\Psi \times \Theta$ such that Theorem 1 ensures the existence of the gradients on $K$. Algorithm \ref{algo:DEPS_algo} summarizes the steps performed in the DEPS algorithm.
\begin{algorithm}[H]
	\caption{DEPS}
	\label{algo:DEPS_algo}
	\begin{algorithmic}
		\Loop
		\State Generate a batch $\mathcal{D}$ of $M$ histories	
		\State Compute the baseline using the histories $B = \frac{1}{m} \sum_{m=0}^{M-1} \sum_{t=0}^{T-1} r_t$
		\State Compute the loss function $\mathcal{L}$ given in equation \eqref{eq:loss_function}
		\State Compute the gradients of the loss function $\mathcal{L}$ via automatic differentiation
		\State Perform gradient ascent as in equation \eqref{eq:grad_update_step}
		\State Project the parameters as in equation \eqref{eq:proj_step}
		\EndLoop
	\end{algorithmic}
\end{algorithm}

The execution of the projected stochastic gradient ascent algorithm for optimizing the objective in equation \eqref{eq:problem_statement} is shown in more details in Algorithm \ref{algo:system_optimization} in Appendix \ref{ap:algos}.

\section{Experiments}\label{sec:experiments}
In this section, we first introduce the methodology used for assessing the performance of DEPS. Afterwards, we test the DEPS algorithm in three parametrized environments and compare its performance with that of alternative approaches. The first environment is a Mass-Spring-Damper (MSD) environment, the second one is related to the design and control of an off-grid microgrid and the last one focuses on the design and control of a drone
\footnote{The implementation of our algorithm and of the different benchmarks are provided in the following GitHub repository: \url{https://github.com/adrienBolland/Jointly-Learning-Environments-and-Control-Policies-with-Projected-Stochastic-Gradient-Ascent}}.

\subsection{Experimental Protocol} \label{sec:experimental_protocol}

In the following, the DEPS algorithm is benchmarked against several methods computing pairs of policies and environments. The performance of each algorithm is measured in terms of the expected return achieved by the associated policy in the associated environment. The expected return is computed from 64 Monte-Carlo samples (i.e., by sampling $64$ i.i.d. trajectories). In addition, we note that the different algorithms we use for selecting policies are stochastic. Hence, we naturally report the average expected return of the pair of policies and environments computed by those algorithms, which is estimated by averaging the performance over ten runs (random seeds) of those algorithms. Finally, the different RL algorithms involve several hyperparameters that have to be tuned to achieve satisfactory results. First, we followed the recommendations of \cite{andrychowicz2020matters} and scaled the state and action variables as well as the parameters that are learned. The scaling factors used for each environment are discussed in sections \ref{sec: msd_exp}, \ref{sec: mg_exp}, and \ref{sec: drone_exp}. Then, the learning rates were tuned in a heuristic way (i.e., we started with a large value and decreased it if the learning did not converge, as suggested by \cite{bengio2012practical}).

First, for each environment, we compare the DEPS algorithm with a state-of-the-art algorithm solving joint design and control problems, namely the so-called JODC algorithm \citep{schaff2019jointly}. The latter is also a gradient-based iterative RL algorithm and is described in detail in Appendix \ref{ap:joint_opt}. For both algorithms, in every iteration $k$, we compute the average expected return of the policy in the environment for the current pair of parameter vectors $(\theta_k, \psi_k)$. In addition, the normalized second-order lower and upper partial moments with respect to the expectation, which will be denoted as $\sigma^-$ and $\sigma^+$, respectively, are also computed with the ten runs of the algorithms and are reported as a confidence band.

Second, we want to verify that the DEPS algorithm indeed converges to a (near-) global optimum. This could, for instance, be done by discretizing the joint space of policy and environment parameters very finely and selecting the optimal pair from the resulting (finite) set of policy-environment pairs. This pair would thus be globally optimal, up to a certain precision. However, this approach is computationally intractable in practice, especially given the high dimension of the space of policy parameters in our benchmarks. We therefore reduce the complexity introduced by the discretization of the space of policy parameters by applying the REINFORCE algorithm to compute a (near-) optimal policy for each set of environment parameters $\psi_d \in \Psi_d$ resulting from the discretization $\Psi_d \subsetneq \Psi$ of the space of environment parameters $\Psi$. The best policy-environment pair is then selected from this reduced set of pairs. Assuming that the REINFORCE algorithm finds a (near-) optimal policy, the method then still yields a (near-) optimal policy-environment pair. In the following, this optimization technique is referred to as the REINFORCE Applied on a Discretized Set of Environments (RADE). Ideally, we would like to apply this method to each of the parametrized environments but this approach is only computationally tractable for the first benchmark (discussed in Section \ref{sec: msd_exp}), where the problem can be reduced to a two-dimensional discretization of the space of environment parameters and where the REINFORCE algorithm has been observed to converge to near-optimal solutions in a few iterations.

Finally, we want to compare the difference in final performance between a joint optimization algorithm, such as DEPS, and an alternative optimization algorithm such that a rule-based policy is fixed \textit{a priori} and the environment is optimized so as to guarantee that the fixed policy has a high expected return. Such policies often take the form of myopic control rules that depend on environment parameters and are designed using expert knowledge. More specifically, we use the \textit{dual annealing} algorithm \citep{xiang2000efficiency}, which is a derivative-free optimization technique, to navigate towards an environment where the policy has a high expected return, as described in Appendix \ref{ap:joint_opt}. In the following, this method is referred to as the rule-based optimization algorithm. This method is not applicable to the third environment that is too complex to develop such a rule-based policy \textit{a priori}.

\subsection{Mass-Spring-Damper Environment} \label{sec: msd_exp}
In section \ref{sec:exp_msd_joind_design}, we briefly describe the Mass-Spring-Damper (MSD) environment and detail the different parameters of the DEPS algorithm. Section \ref{sec:exp_msd_results} discusses simulation results.

\subsubsection{Parameters of the Joint Design and Control Problem} \label{sec:exp_msd_joind_design}
Let us describe the different elements of the optimization problem we are trying to solve when jointly designing and controlling the MSD.

\paragraph{Parametrized environment.} We consider the MSD environment where we model the motion of an object attached to a spring and a damper that we aim to stabilize at a reference position $x_{ref}$. The environment is described in detail in Appendix \ref{ap:msd}.

\paragraph{Hypothesis Spaces.} The environment is parametrized by the real vector $\psi \in \Psi$, where $\psi = (\omega, \zeta, \phi_0, \phi_1, \phi_2) \in \Psi = [0.1,\: 1.5] \times [0.1,\: 1.5] \times [-2,\: 2] \times [-2,\: 2] \times [-2,\: 2]$. The hypothesis space of policy parameters $\Theta$ is defined as follows. Any policy in the parameter space is a multi-layer perceptron (MLP) with three inputs: one for each of the $|\St|=2$ values of the state vector $s_t$ and one for the time $t$. Each MLP is composed of one hidden layer featuring $64$ neurons with hyperbolic tangent activation functions and has five output neurons ($|\A| = 5$) without activation and from which a probability distribution over $\A$ is inferred using a softmax function.

\paragraph{Parameters of the DEPS Algorithm.} The gradients are evaluated by applying automatic differentiation on the loss function defined in equation \eqref{eq:loss_function}. Furthermore, the \textit{Adam} algorithm \citep{kingma2014adam} is used for updating $(\psi, \theta)$. This algorithm is a variant of the vanilla stochastic gradient ascent given in Algorithm \ref{algo:system_optimization} that has proven to perform well on highly non-convex problems. The gradients are estimated on batches of $M = 64$ trajectories and the step size $\alpha$ of the Adam algorithm is chosen equal to $0.005$ for both the environment and the policy gradients. We retain the default values for the other parameters of the Adam algorithm. Moreover, the inputs of the policy are z-normalized using mean vector $(x_{ref}, 0, 0)$ and standard deviation vector $(0.005, 0.02, 100)$, which is an approximation of the standard deviation vector of the states collected over high-performing trajectories.

\subsubsection{Experimental Results}
\label{sec:exp_msd_results}
In this section, we apply the experimental protocol of Section \ref{sec:experimental_protocol} to the MSD environment and provide an analysis of the experimental results.

\begin{figure}[h]
	\centering
	\begin{subfigure}[t]{0.5\textwidth}
		\centering
		\includegraphics[width=0.85\linewidth]{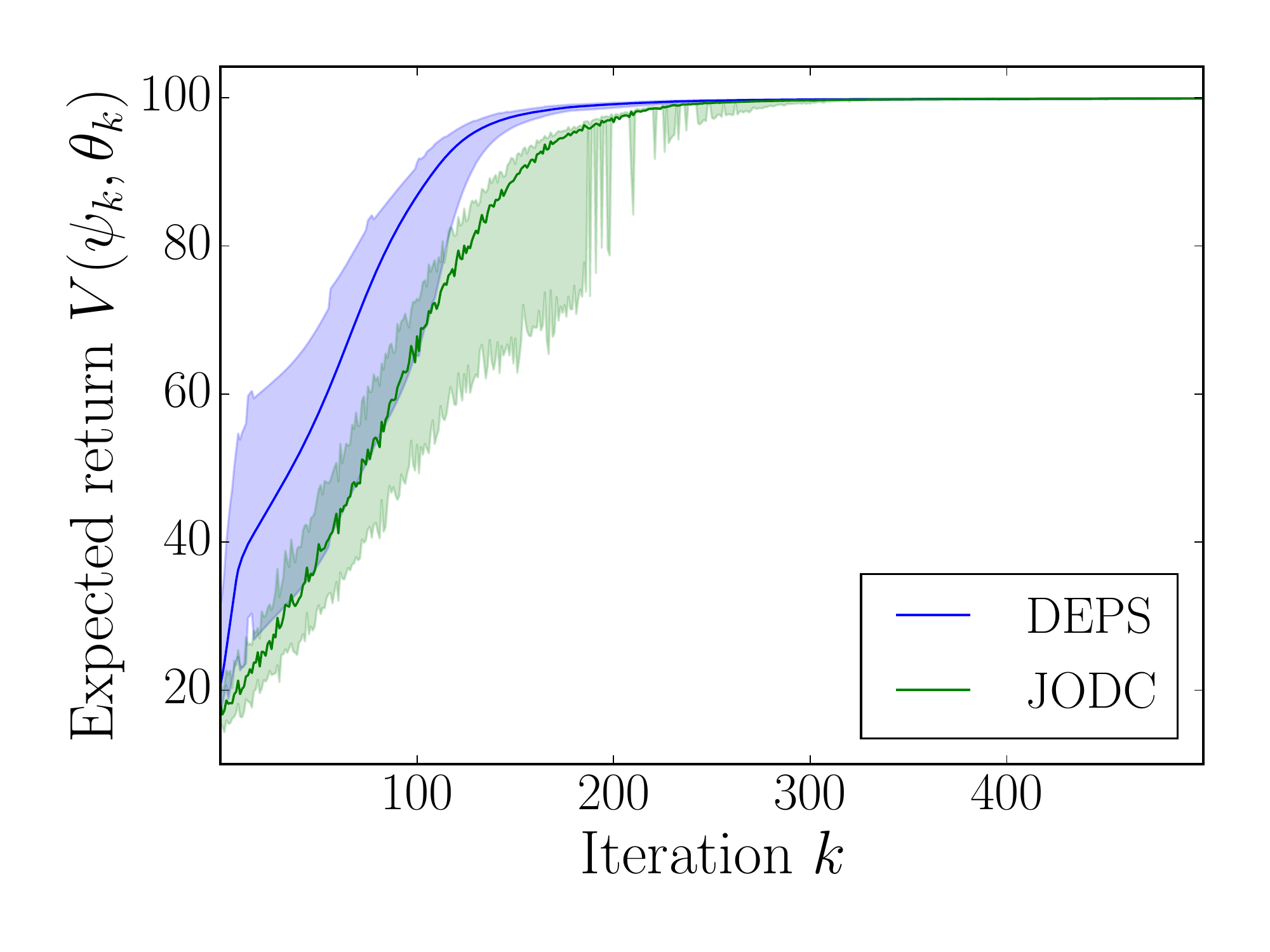}
		\caption{Batches of $M=64$ trajectories.}
		\label{fig:exp_avg_perf_mdp}
	\end{subfigure}%
	~ 
	\begin{subfigure}[t]{0.5\textwidth}
		\centering
		\includegraphics[width=0.85\linewidth]{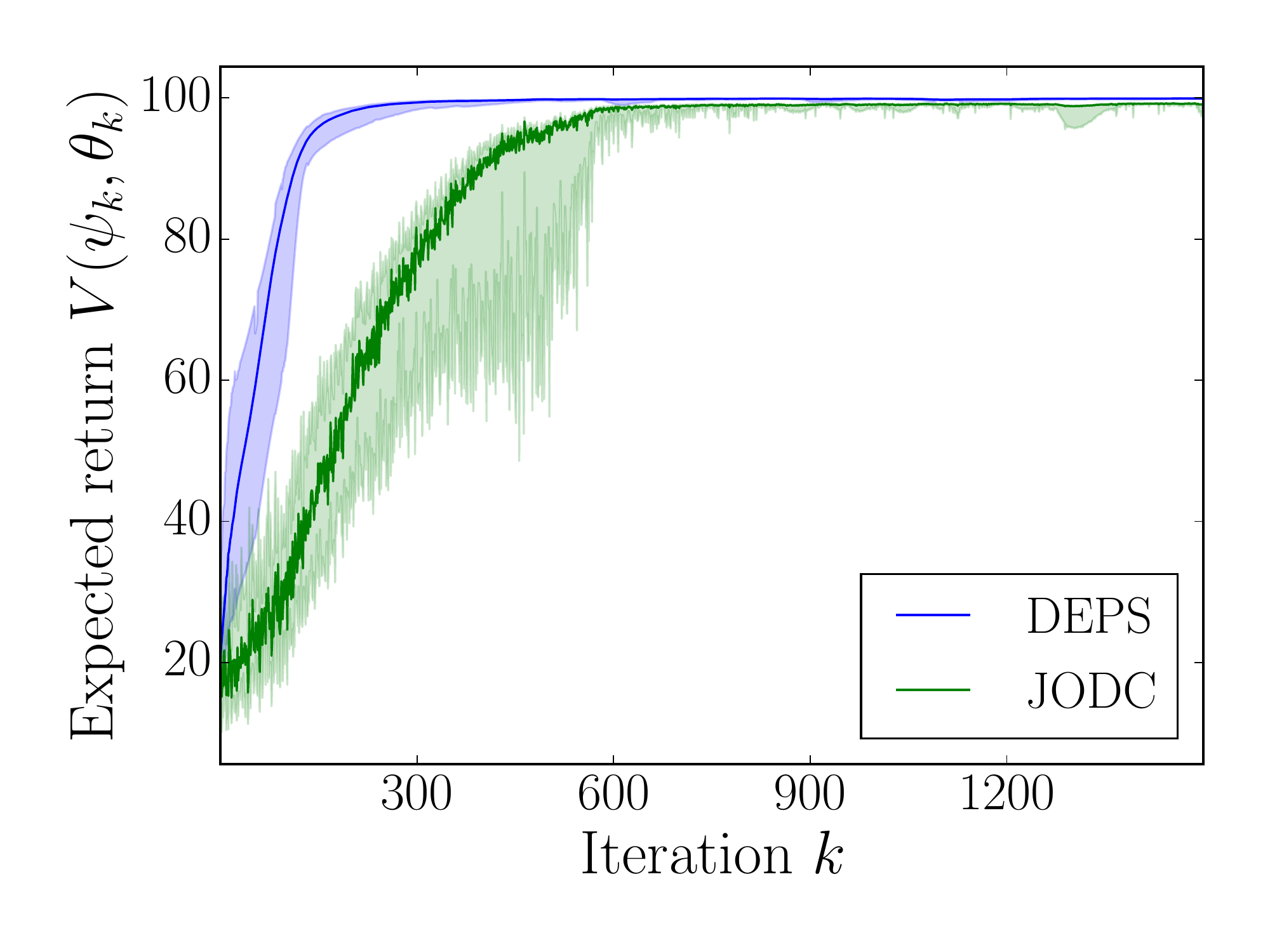}
		\caption{Batches of $M=4$ trajectories.}
		\label{fig:exp_avg_perf_2_mdp}
	\end{subfigure}
	\caption{Average expected return of the policy in the environment parametrized by $(\theta_k, \psi_k)$ as a function of the iteration number $k$ of the DEPS and the JODC algorithms when using batches of $M=64$ trajectories (left) and batches of $M=4$ trajectories (right) in the MSD environment.}
\end{figure}

\paragraph{Performance of DEPS.} The blue line in Figure \ref{fig:exp_avg_perf_mdp} shows the evolution of the average expected return of the DEPS algorithm as the iteration count grows. The normalized partial moments $\sigma^-$ and $\sigma^+$ between the different runs are illustrated by the shaded area under and above the average, respectively. The values of these statistics for the final pairs of policy and environment are reported in Table \ref{tab:msd-return-val}. As we can see, the DEPS algorithm converges towards a maximal expected return almost equal to $100$. We note that $100$ is an upper-bound on the return that can only be reached if at each time step $t$, the position of the mass is at the reference position $x_{ref}$, as detailed in Appendix \ref{ap:msd}. The moments $\sigma^-$ and $\sigma^+$ also strongly decrease as the iteration count increases. 
Table \ref{tab:msd-psi-param} provides the average and the standard deviation of the final value of the parameter $\psi$, computed over the ten runs of the algorithm. For each simulation, the algorithm converges to a parameter $\psi = (\omega,\zeta, \phi_0, \phi_1, \phi_2)$, where $\omega$ and $\zeta$ are both equal to $0.5$ and where (at least) one of the components $\phi_0$, $\phi_1$ or $ \phi_2$ is equal to $c_0$, $c_1$ or $c_2$, respectively. We note that any triplet $(\phi_0, \phi_1, \phi_2)$ satisfying this condition can be considered optimal, as described in Appendix \ref{ap:msd}.

\paragraph{Comparison with JODC.} In Figure \ref{fig:exp_avg_perf_mdp}, the evolution of the average expected return of the JODC algorithm (using the parameters given in Appendix \ref{ap:optim_parameters}) is also reported in green. In addition, Table \ref{tab:msd-return-val} provides the average and the moments of the expected return for the final pairs of policy and environment. As can be seen in the table, both algorithms converge to parameters leading to similar average expected returns. However, JODC converges more slowly and its variance is higher across the different simulations, as can be seen in Figure \ref{fig:exp_avg_perf_mdp}. This phenomenon can also be observed in Figure \ref{fig:exp_avg_perf_2_mdp}, where the same experiments were carried out with batches of $M=4$ trajectories. These results suggest that the DEPS algorithm is less subject to noise than the JODC algorithm and is thus more stable. Furthermore, DEPS manages to reach better performance across the different simulations faster than the JODC algorithm, indicating that DEPS is also more sample-efficient. Concerning the parameters of the environment, the JODC algorithm converges to values that are similar to the ones identified by the DEPS algorithm, as can be seen from Table \ref{tab:msd-psi-param}.

\begin{figure}[H]
	\centering
	\begin{subfigure}[t]{0.5\textwidth}
		\centering
		\includegraphics[width=1.0\linewidth]{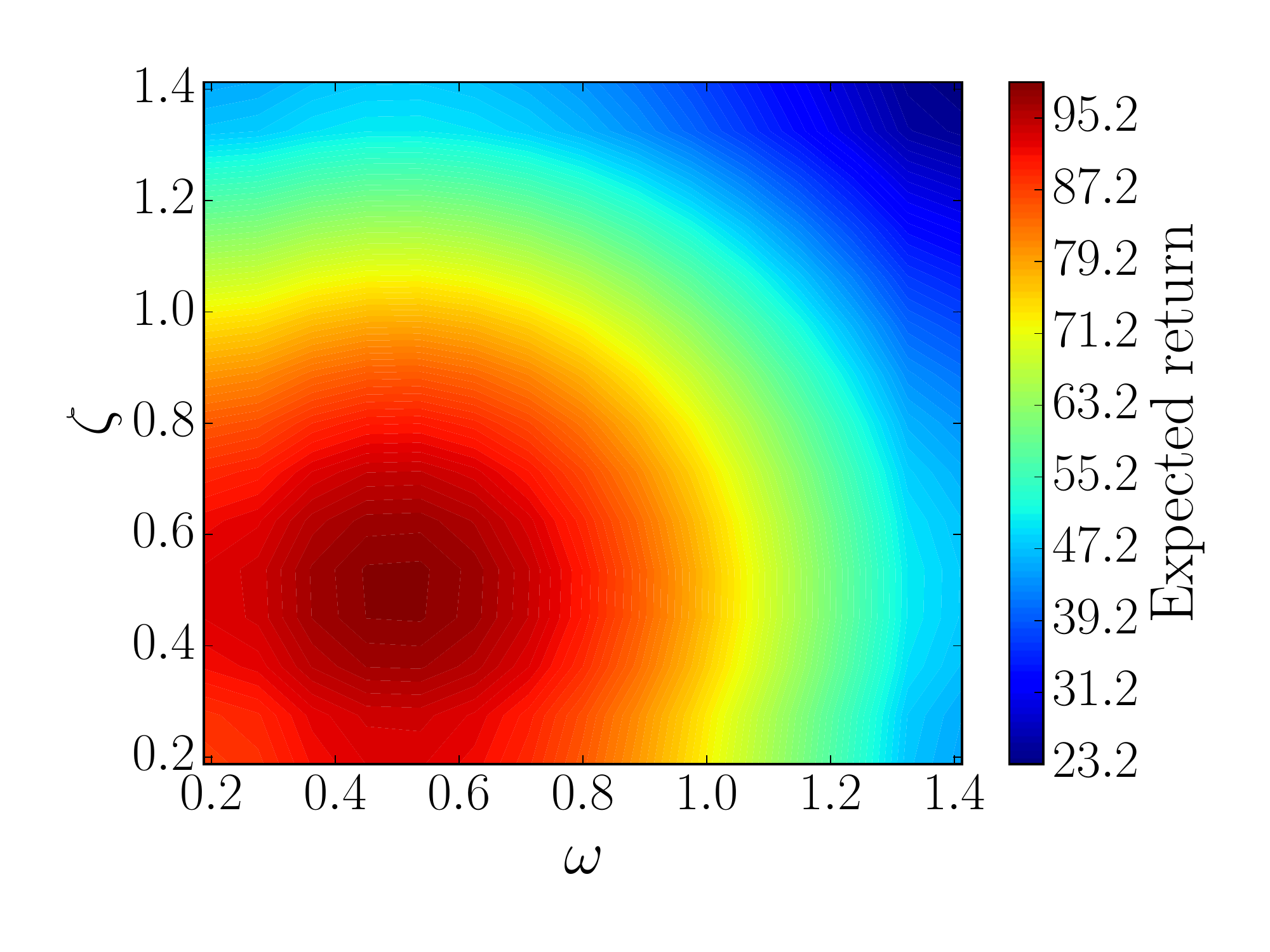}
	\end{subfigure}
	\caption{Average expected return of each pair of policy and environment computed with the RADE algorithm in the MSD environment.}
	\label{fig:exp_grid_search_mdp}
\end{figure}

\paragraph{Comparison with RADE.} Figure \ref{fig:exp_grid_search_mdp} shows the average expected return of each pair of environment and policy computed with the RADE algorithm, where $\psi_d \in \Psi_d = \Omega_d \times Z_d \times \{c_0\} \times \{c_1\} \times \{c_2\}$ and where $\Omega_d = Z_d = \{0.1 + k\cdot \Delta|k=1, \dots, 15\}$ with $\Delta = 0.082$. We note that $c_0$, $c_1$ and $c_2$ correspond to a triplet of optimal values for $\phi_1$, $\phi_2$ and $\phi_3$, respectively, as described in Appendix \ref{ap:msd}. Similarly to the DEPS and the JODC algorithms, the highest average expected return of the policies occurs for $(\omega, \zeta) = (0.5, 0.5)$. In addition, the highest average return of the policies identified by the REINFORCE algorithm was almost identical to the average expected returns obtained by the aforementioned algorithms. This experiment indicates that, under the previously described assumptions on the convergence of the RADE algorithm, the DEPS algorithm converges to an optimal policy-environment pair.

\paragraph{Comparison with Rule-Based Optimization.} When applying the rule-based optimization method with policies $\pi_{msd, 1}$ and $\pi_{msd, 2}$ (which are detailed in Appendix \ref{ap:msd}), it takes an average of $2251$ and $2554$ iterations of the dual annealing algorithm \citep{xiang2000efficiency} to identify an optimal set of environment parameters. The standard deviations are equal to $239$ and $409$ iterations for $\pi_{msd, 1}$ and $\pi_{msd, 2}$, respectively. Table \ref{tab:msd-return-val} provides the average value and the partial moments of the expected return of the computed pairs of policy and environment parameters. Table \ref{tab:msd-psi-param} provides the average and the standard deviation of the parameters $\psi$ computed by the rule-based optimization method. For the first policy $\pi_{msd, 1}$, the resulting average expected return is lower than the average returns obtained by the DEPS and the JODC algorithms. The second policy $\pi_{msd, 2}$, on the other hand, achieved average returns that are similar to the ones of DEPS and JODC. Let us note that the optimization with $\pi_{msd, 1}$ overestimates the optimal pair of parameters $(\omega, \zeta)$ while the optimization with $\pi_{msd, 2}$ converges towards a near-optimal pair. This experiment allows us to conclude that the performance of the method is strongly related to the quality of the rule-based policy that is used. Moreover, even in the case where a policy reflecting expert knowledge about the environment is used, the final average expected return is lower than that achieved with the joint optimization of the policy and the environment. This validates the hypothesis that the joint optimization of environment and control policy offers a real advantage in terms of final performance.
\begin{table}[H]
	\begin{center}
		\renewcommand\arraystretch{1}
		\caption{Final expected return in the MSD environment.}
		\begin{tabular}[b]{l c c c}
			\hline
			Algorithm & Average value & Moment $\sigma^-$ & Moment $\sigma^+$ \\
			\hline
			DEPS & $99.90$ & $0.06$ & $0.04$ \\
			JODC & $99.80$ & $1.18$ & $0.13$ \\
			Rule-based $\pi_{msd, 1}$ & $62.93$ & $0.29$ & $0.26$ \\
			Rule-based $\pi_{msd, 2}$ & $99.55$ & $0.05$ & $0.03$ \\
			\hline
		\end{tabular}
		\label{tab:msd-return-val}
	\end{center}
\end{table}

\begin{table}[H]
	\begin{center}
		\renewcommand\arraystretch{1}
		\caption{Final parameters computed for the MSD environment.}
		\begin{tabular}[b]{c r r r r r r}
			\hline
			Algorithm & \makecell[c]{$\omega$} & \makecell[c]{$\zeta$} & \makecell[c]{$\phi_0$} & \makecell[c]{$\phi_1$} & \makecell[c]{$\phi_2$} \\
			\hline
			DEPS & $0.50 \pm 0.01$ &  $0.50 \pm 0.00$ & $0.45 \pm 0.51$ & $ -0.17 \pm 0.63$ & $0.16 \pm 0.82$ \\
			JODC & $0.50 \pm 0.01$ & $0.50 \pm 0.01$ & $0.54 \pm 0.38$ & $-0.32 \pm 0.83$ & $0.39 \pm 1.11$ \\
			Rule-based $\pi_{msd, 1}$ & $0.57 \pm 0.03$ & $0.70 \pm 0.03$ & $0.16 \pm 0.95$ & $-0.70 \pm 0.68$ & $0.05 \pm 0.46$ \\
			Rule-based $\pi_{msd, 2}$ & $0.50 \pm 0.01$ & $0.50 \pm 0.01$ & $-0.17 \pm 0.99$ & $-0.22 \pm 0.53$ & $-0.04 \pm 0.40$ \\
			\hline
		\end{tabular}
		\label{tab:msd-psi-param}
	\end{center}
\end{table}

\subsection{Sizing and Operation of an Off-Grid Microgrid} \label{sec: mg_exp}
In section \ref{sec:exp_mg_joind_design}, we briefly describe the microgrid environment and detail the different parameters of the DEPS algorithm. Section \ref{sec:exp_mg_results} discusses simulation results.

\subsubsection{Parameters of the Joint Design and Control Problem} \label{sec:exp_mg_joind_design}
Let us describe the different elements of the joint design and control optimization problem related to the microgrid environment.

\paragraph{Parametrized environment.} We consider an environment modeling a problem where an agent seeks to design a small-scale off-grid electrical power system and operate it so as to minimize the cost of serving some electricity demand (including investment and operating costs) over the system lifetime. The investment costs are proportional to the size of the components of the microgrid, namely the battery, the solar panels and the diesel generator. Operating costs stem from any mismatch that may arise between production and consumption in the system as well as from the operation of the diesel generator. The environment is presented in Appendix \ref{ap:mg}.

\paragraph{Hypothesis Spaces.} The environment is parametrized by the vector $\psi = (C^B, C^{PV}, C^{G}) \in \Psi$, where the parameter space is chosen equal to $\Psi = [20, 200] \times [20, 200] \times [1.6, 16]$. The selection of parameter ranges is based on their respective physical interpretations. In particular, simultaneously setting all parameters equal to their upper and lower bounds yields large and small installations, respectively. More specifically, for the larger installation, the daily solar production nearly equals the daily consumption level. In this context, the full daily PV production can be stored in the battery. Finally, it is possible to cover the whole hourly consumption with the diesel generator. For the smaller installation, the consumption level is impossible to cover which results in high unserved demand penalties. We will constrain the hypothesis space for the policies to multivariate Gaussian policies of dimension $|\A| = 2$ such that $\pi_\theta(a_t|s_t, t) = \mathcal{N}(a_t| \mu_\theta(s_t, t), I \cdot \sigma_\theta^2(s_t, t)), \: \forall a_t \in \A, \forall s_t \in \St, \forall t \in \{0, \dots, T-1\}$ where $I \in \mathbb{R}^2$ is the identity matrix of size two and where $\mu_\theta(s_t, t) \in \mathbb{R}^2$ and $\sigma_\theta(s_t, t) \in \mathbb{R}^2$ are the expectation and the standard deviation of the multivariate normal distribution $\mathcal{N}$. The expectation and standard deviation are expressed as functions of the state $s_t$, the time $t$ and the parameter vector $\theta$. Let us note that the actions are mutually independent (i.e., their covariance is equal to zero). The vectors $\mu_\theta(s_t, t)$ and $\sigma_\theta^2(s_t, t)$ are output by an MLP taking seven values as input: one for each of the $|\St| = 6$ values of the state vector $s_t$ and one for the time $t$. The MLP is composed of one hidden layer of $64$ neurons with hyperbolic tangent activation functions and has four output neurons. The two outputs corresponding to $\mu_\theta(s_t, t)$ do not have activation functions while the outputs corresponding to $\sigma_\theta^2(s_t, t)$ are squared by their activation function. The set of values that the parameters of the MLP can take defines the space of policy parameters $\Theta$.

\paragraph{Parameters of the DEPS Algorithm.} Similarly to the MSD environment, the gradients are evaluated using automatic differentiation and the parameters are updated with the Adam algorithm \citep{kingma2014adam}. The gradients are estimated on batches of $M = 64$ trajectories and the step size $\alpha$ of the Adam algorithm is chosen equal to $0.001$ for both the environment and the policy parameters. The default values are kept for the other parameters of the Adam algorithm. In addition, the norm of the gradient of the transition function in equation \eqref{eq:grad_s} is scaled in order to avoid gradient explosion (here, the scaling guarantees that the norm of gradients does not exceed one hundred billion). Doing so keeps the direction of the gradient unchanged and avoid numerical issues when performing gradient ascent. The input of the MLP corresponding to the time $t$ is scaled by the horizon $T=120$. Furthermore, the rewards collected are scaled linearly from the interval $\left[ -5000,0\right] $ to the interval $\left[ 0,1\right]$. Finally, the vector $\psi$ is scaled by the vector $(100, 100, 8)$ in order to keep the order of magnitude of the (scaled) environment parameters similar to that of the MLP parameters.

\subsubsection{Experimental Results}
\label{sec:exp_mg_results}
In this section, we apply the experimental protocol of Section \ref{sec:experimental_protocol} to the microgrid environment and provide an analysis of the experimental results.

\begin{figure}[H]
	\centering
	\begin{subfigure}[t]{0.5\textwidth}
		\centering
		\includegraphics[width=1.0\linewidth]{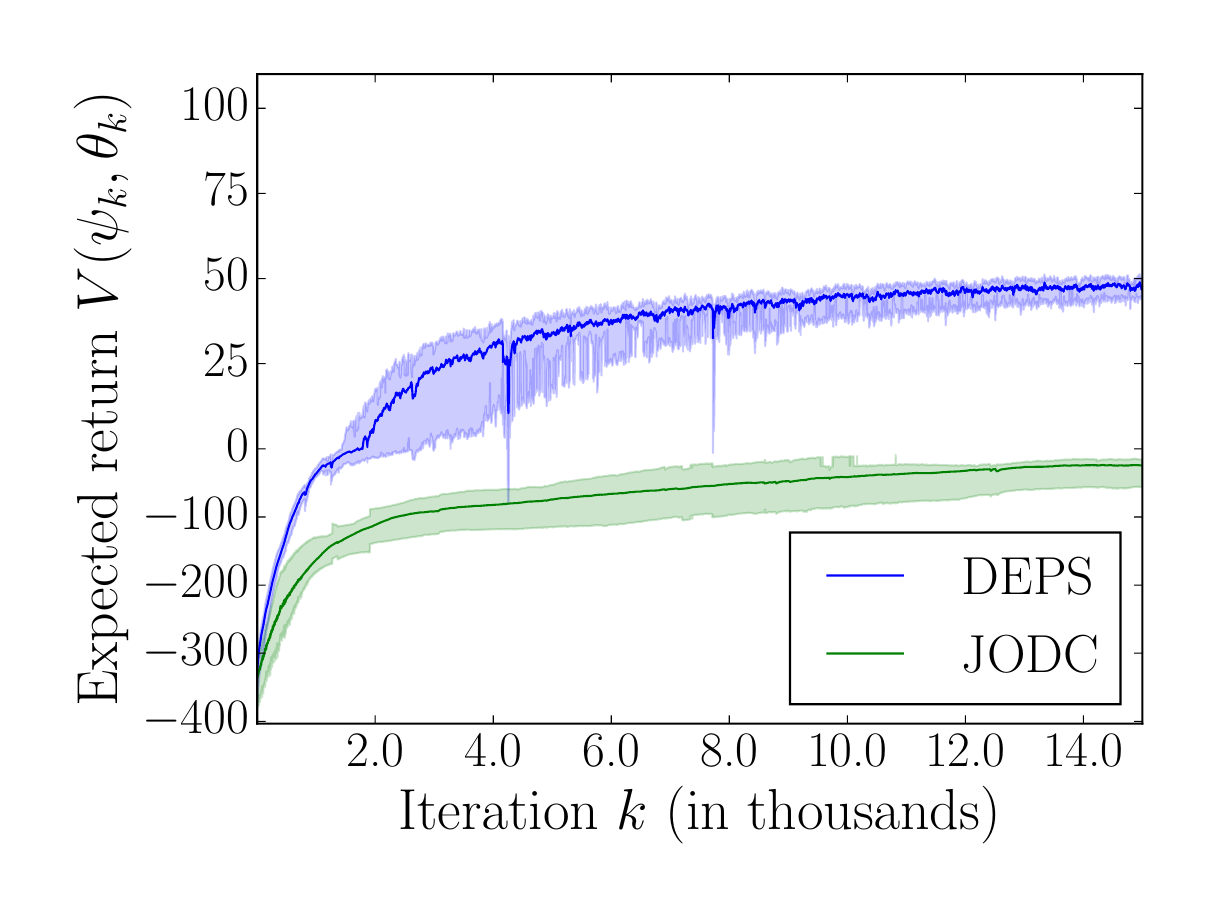}
	\end{subfigure}%
	\caption{Average expected return of the policy in the environment parametrized by $(\theta_k, \psi_k)$ as a function of the iteration number $k$ of the DEPS and JODC algorithms  in the microgrid environment.}
	\label{fig:exp_avg_perf_mg}
\end{figure}

\paragraph{Performance of DEPS.} The blue line in Figure \ref{fig:exp_avg_perf_mg} presents the evolution of the average expected (scaled) return collected in the off-grid microgrid environment when applying the DEPS algorithm. It can be seen in Figure \ref{fig:exp_avg_perf_mg} that the moments $\sigma^-$ and $\sigma^+$ strongly decrease as the iteration count increases. The final average expected return and moments are reported in Table \ref{tab:mg-return-val}. In addition, the average and the standard deviation of the environment parameters learned are reported in Table \ref{tab:mg-psi-param}. Let us also note that in this experiment, the parameters $\psi$ learned by the algorithm suffer from a rather high degree of variance, indicating that several combinations of parameters yield high expected returns.

\paragraph{Comparison with JODC.} The evolution of the average expected (scaled) return collected when applying the JODC algorithm is also reported in Figure \ref{fig:exp_avg_perf_mg}. Table \ref{tab:mg-return-val} provides the final value. As can be seen from Figure \ref{fig:exp_avg_perf_mg}, the JODC algorithm converges to policy-environment pairs with a lower average expected return compared with the DEPS algorithm. This difference could be partly explained by the fact that the JODC algorithm converges to parameters $\psi$ that differ on average from the parameters found by DEPS and have a much higher degree of variance, as shown in Table \ref{tab:mg-psi-param}.

\paragraph{Comparison with Rule-Based Optimization.} 

When applying the rule-based optimization method with policies $\pi_{mg, 1}$ and $\pi_{mg, 2}$ (which are detailed in Appendix \ref{ap:mg}), it takes an average of $4967$ and $4931$ iterations of the dual annealing algorithm \citep{xiang2000efficiency} to identify an optimal set of environment parameters. The standard deviations are equal to $649$ and $455$ iterations for $\pi_{mg, 1}$ and $\pi_{mg, 2}$, respectively. Table \ref{tab:mg-return-val} provides the average value and the partial moments of the expected return of the computed pairs of policy and environment parameters. Table \ref{tab:mg-psi-param} provides the average and the standard deviation of the parameters $\psi$ computed by the rule-based optimization method. Both optimization procedures result in final average expected returns that are slightly lower than that of the DEPS algorithm, which supports the conclusions already drawn in Section \ref{sec: msd_exp}. In particular, we see that different policies result in very different environment configurations, which emphasizes the interdependence between environment and policy parameters and thus suggests that both should be jointly optimized to achieve high expected returns.

\begin{table}[H]
	\begin{center}
		\renewcommand\arraystretch{1}
		\caption{Final expected return in the microgrid environment.}
		\begin{tabular}[b]{l c c c}
			\hline
			Algorithm & Average value & Moment $\sigma^-$ & Moment $\sigma^+$ \\
			\hline
			DEPS & $47.60$ & $5.79$ & $4.97$ \\
			JODC &  $-24.17$ & $53.71$ & $65.53$ \\
			Rule-based $\pi_{mg, 1}$ & $44.08$ & $0.29$ & $0.27$ \\
			Rule-based $\pi_{mg, 2}$ & $45.90$ & $1.62$ & $0.85$ \\
			\hline
		\end{tabular}
		\label{tab:mg-return-val}
	\end{center}
\end{table}

\begin{table}[H]
	\begin{center}
		\renewcommand\arraystretch{1}
		\caption{Final environment parameters computed for the microgrid environment.}
		\begin{tabular}[b]{l r r r}
			\hline
			Algorithm & \makecell[c]{$C^B$}  & \makecell[c]{$C^{PV}$}  & \makecell[c]{$C^{G}$}  \\
			\hline
			DEPS & $131.23 \pm 30.03$ & $73.57 \pm 12.18$ & $5.96 \pm 0.88$ \\
			JODC & $102.50 \pm 30.88$ & $34.97 \pm 39.21$ & $6.86 \pm 2.23$ \\
			Rule-based $\pi_{mg, 1}$ & $164.73 \pm 5.09$ & $86.52 \pm 4.58$ & $8.74 \pm 0.18$ \\
			Rule-based $\pi_{mg, 2}$ & $127.80 \pm 26.56$ & $95.73 \pm 14.07$ & $4.64 \pm 1.02$ \\
			\hline
		\end{tabular}
		\label{tab:mg-psi-param}
	\end{center}
\end{table}

\subsection{Drone Design and Control} \label{sec: drone_exp}
In section \ref{sec:exp_drone_joind_design}, we briefly describe the drone environment. Section \ref{sec:exp_drone_results} discusses simulation results.

\subsubsection{Parameters of the Joint Design and Control Problem} \label{sec:exp_drone_joind_design}
Let us describe the different elements of the joint design and control optimization problem related to the drone environment.

\paragraph{Parametrized environment.} In this section, we consider the drone environment where we design and control a drone that has to fly along an elliptical trajectory as fast as possible. This environment is presented in detail in Appendix \ref{ap:drone}.

\paragraph{Hypothesis Spaces.} The environment is parametrized by the vector $\psi = (D, R, H, W) \in \Psi$, where the parameter space is chosen equal to $\Psi = [0.05, 0.2] \times [0.01, 0.2] \times [0.001, 0.01] \times [0.001, 0.01]$ and represents realistic ranges for the geometric parameters defining the drone design. The environment is parametrized by the real vector $\psi \in \Psi$, where $\psi = (D, R, H, W) \in \Psi = [0.05, 0.2] \times [0.01, 0.2] \times [0.001, 0.01] \times [0.001, 0.01]$. Similar to the hypothesis space of policies in the microgrid environment, multivariate Gaussian policies of dimension $|\A| = 4$ of the same form as in Section \ref{sec: mg_exp} are considered. The expectation $\mu_\theta(s_t, t) \in \mathbb{R}^4$ and the variance $\sigma_\theta^2(s_t, t) \in \mathbb{R}^4$ of the distribution are provided by an MLP. The network takes as input the $|\St| = 12$ state variables and the time $t$, has one hidden layer of $64$ neurons using hyperbolic tangent activation functions, and has eight output neurons with parabolic activation functions for the outputs corresponding to $\sigma_\theta^2(s_t, t)$ and without activation for the outputs corresponding to $\mu_\theta(s_t, t)$. The space of policy parameters $\Theta$ is defined by the set of values that the parameters of the MLP may take.
        
\paragraph{Parameters of the DEPS Algorithm.} Similar to the MSD and the microgrid environments, the gradients are evaluated using automatic differentiation and the parameters are updated with the Adam algorithm \citep{kingma2014adam}. The same batch size of $M = 64$ trajectories and the same scaling strategy for the gradient of the transition function as in Section \ref{sec: mg_exp} are used. In addition, the step size $\alpha$ of the Adam algorithm is chosen equal to $0.00005$ for the gradient ascent steps taken in both the environment and policy spaces. Moreover, \cite{andrychowicz2020matters} observed that reducing the variance of the initial policy as well as its initial dependency on the observations has a high impact on the training of this policy. \cite{andrychowicz2020matters} recommend initializing the MLP with smaller weights in the last layer in order to achieve this reduction of variance. To this end, we divide the initial weights of the last layer of the MLP by a factor $30$, which yielded the best empirical results, before performing gradient ascent. The input of the MLP is also z-normalized by the mean vector $(0, 0, 0, 0, 0, 0, 0, 0, 0, 1, 0, 0, 0)$ and by the standard deviation vector $(0.01, 0.01, 0.01, 0.01, 0.01, 0.01, 0.1, 0.1, 0.1, 1, 1, 0.1, 100)$. In addition, the states are clipped to stay in the range $[-100, 100]$ during the learning process in order to avoid numerical problems when the policy still behaves sub-optimally and the position of the drone tends to diverge from the desired trajectory. The outputs of the neural network corresponding to the expectation of the distribution $\mathcal{N}$ are centered at the stationary speed required for counterbalancing the mass of the drone. This is achieved by adding $\omega_{stat} = \sqrt{m g / (4 b)}$ to these outputs. Doing so, we only learn the variations in speed of the propellers that are required for moving the drone along the desired trajectory. Finally, the vector $\psi$ is scaled by the vector $(0.2, 0.2, 0.01, 0.01)$ to keep these parameters in a range similar to that of the MLP parameters.

\subsubsection{Experimental Results}
\label{sec:exp_drone_results}
In this section, we apply the experimental protocol of Section \ref{sec:experimental_protocol} to the drone environment and provide an analysis of the experimental results.

\begin{figure}[H]
	\centering
	\begin{subfigure}[t]{0.5\textwidth}
		\centering
		\includegraphics[width=1.0\linewidth]{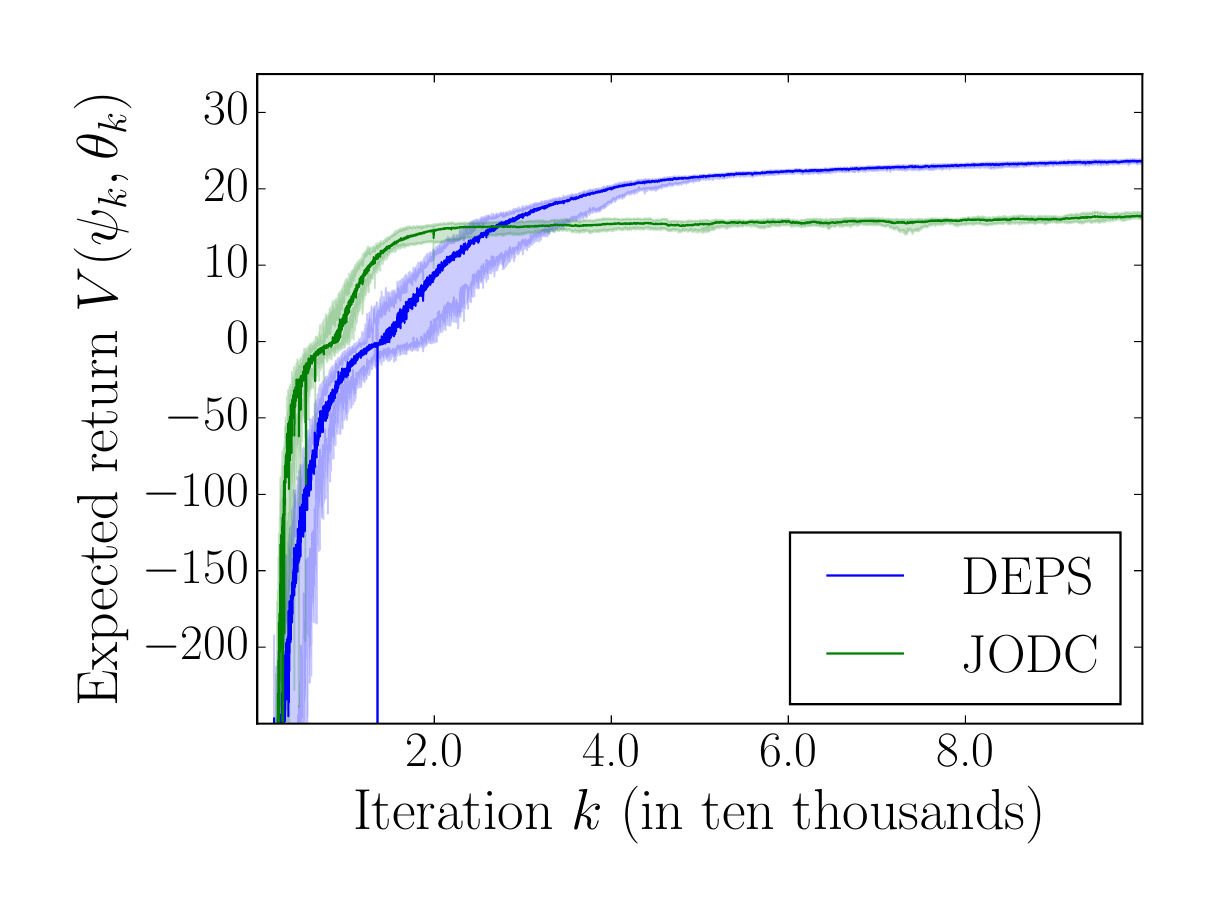}
	\end{subfigure}%
	\caption{Average expected return of the policy in the environment parametrized by $(\theta_k, \psi_k)$ as a function of the iteration number $k$ of the DEPS and the JODC algorithms in the drone environment.}
	\label{fig:exp_avg_perf_drone}
\end{figure}

\paragraph{Performance of DEPS.} The blue line in Figure \ref{fig:exp_avg_perf_drone} presents the evolution of the average expected return collected in the drone environment when applying the DEPS algorithm. 
This value, as well as the partial moments of the final pairs of policy and environment parameters, are reported in Table \ref{tab:drone-return-val}. The average and the standard deviation of the environment parameters computed by DEPS are given in Table \ref{tab:drone-psi-param}. 

\paragraph{Comparison with JODC.} Figure \ref{fig:exp_avg_perf_drone} also provides the evolution of the average expected return collected in the drone environment when applying the JODC algorithm with the parameters given in Appendix \ref{ap:optim_parameters}. As can be seen from Table \ref{tab:drone-return-val}, the DEPS algorithm finds a better combination of parameters compared to the JODC algorithm. The average and the standard deviation of the parameters $\psi$ are given in Table \ref{tab:drone-psi-param}. Unlike the DEPS algorithm, the JODC algorithm never converges to a fixed environment over its ten runs. Instead, the parameters identified by JODC oscillate throughout the iterations of the algorithm around the set of parameter values $\psi = (0.05, 0.2, 0.01, 0.01)$ identified by DEPS. This observation partly accounts for the difference between the final average expected returns produced by the two algorithms.

\begin{table}[H]
	\begin{center}
		\renewcommand\arraystretch{1}
		\caption{Final expected return in the drone environment.}
		\begin{tabular}[b]{l c c c}
			\hline
			Algorithm & Average value & Moment $\sigma^-$ & Moment $\sigma^+$ \\
			\hline
			DEPS & $23.20$ & $0.30$ & $0.30$ \\
			JODC & $16.39$ & $1.23$ & -$0.65$ \\
			\hline
		\end{tabular}
		\label{tab:drone-return-val}
	\end{center}
\end{table}

\begin{table}[H]
	\begin{center}
		\renewcommand\arraystretch{1}
		\caption{Final parameters computed for the drone environment.}
		\begin{tabular}[b]{l r r r r}
			\hline
			Algorithm & \makecell[c]{$D$} & \makecell[c]{$R$} & \makecell[c]{$H$} & \makecell[c]{$W$} \\
			\hline
			DEPS & $0.05 \pm 0.00$ & $0.02 \pm 0.00$ & $0.01 \pm 0.00$ & $0.01 \pm 0.00$ \\
			JODC & $0.05 \pm 0.07$ & $0.02 \pm 0.03$ & $0.01 \pm 0.00$ & $0.01 \pm 0.00$ \\
			\hline
		\end{tabular}
		\label{tab:drone-psi-param}
	\end{center}
\end{table}

\begin{figure}[H]
	\centering
	\begin{subfigure}[t]{0.5\textwidth}
		\centering
		\includegraphics[width=1.0\linewidth]{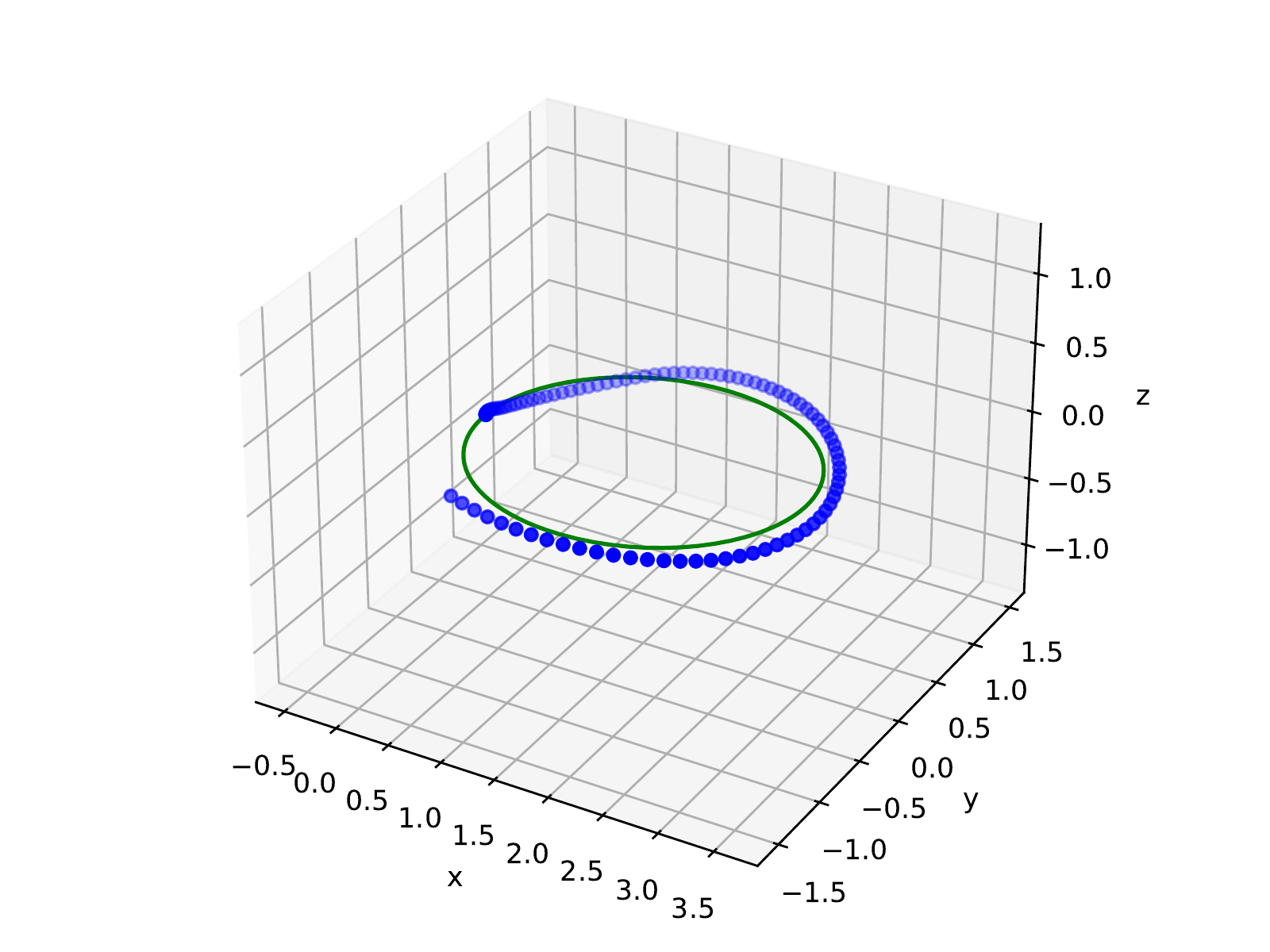}
		\caption{Drone trajectory obtained with DEPS.}
	\end{subfigure}%
	~ 
	\begin{subfigure}[t]{0.5\textwidth}
		\centering
		\includegraphics[width=1.0\linewidth]{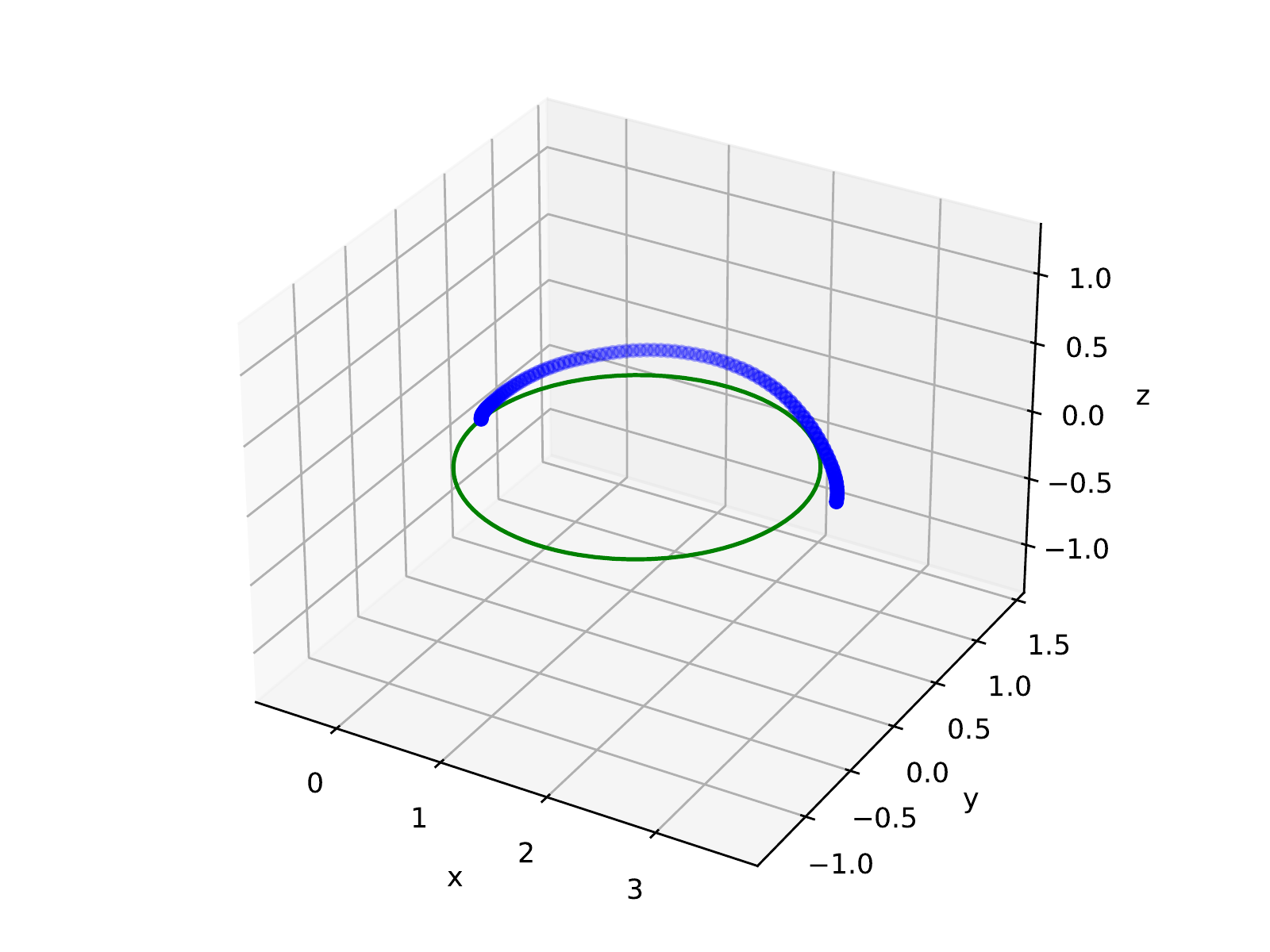}
		\caption{Drone trajectory obtained with JODC.}
	\end{subfigure}
	\caption{Typical trajectory obtained by applying a control policy in an environment optimized with DEPS (left) and with JODC (right). The elliptical trajectory that the drone should follow is shown in green, while the successive positions traveled clockwise by the drone are shown in blue.}
	\label{fig:traj_drone}
\end{figure}

Since constructing a meaningful rule-based policy \textit{a priori} or applying the RADE algorithm is particularly challenging for the drone environment, we have no other points of comparison to evaluate the performance of the DEPS and JODC algorithms. We thus validate the quality of the policy and the environment learned by both algorithms graphically. Figure \ref{fig:traj_drone} shows two typical drone trajectories sampled from the policies and environments learned with the DEPS and JODC algorithms. Both algorithms manage to learn environment and policy parameters resulting in a drone that follows the elliptic trajectory. In the case of JODC, the drone travels over a smaller portion of the ellipse compared with DEPS. Its average speed is thus lower. This phenomenon explains the fact that the DEPS algorithm performs better. We also previously stated that both algorithms converge (or approach, in the case of JODC) to parameters $\psi = (0.05, 0.2, 0.01, 0.01)$. Bearing in mind that these parameters essentially define the mass and moments of inertia of the drone, the drone design resulting from the optimization process in fact has high translational inertia and low rotational inertia around its principal axes. This observation makes direct physical sense, as a heavy drone is less sensitive to wind gusts and a drone with low moments of inertia is more maneuverable.

\subsection{Discussion}
In this section, we discuss the two approaches that we have tested to perform joint design and control, namely DEPS and JODC. We first compare the hypotheses required for computing a solution to the problem statement using the DEPS and the JODC algorithms (i.e., their applicability to variants of the problem). Then, we summarize the differences observed between these algorithms in the previous experiments. More particularly, we focus on the speed of convergence of the algorithms and on the quality of the final solution.

\paragraph{Applicability.} As already discussed in Section \ref{sec:gradient_computation}, DEPS is applicable under smoothness assumptions imposing restrictions on the transition function, on the reward function, as well as on the disturbance probability distribution of the dynamical system. In practice, it implies that the (first-order) derivatives of these functions must exist and be readily computable in order to identify a policy-environment pair. Contrary to this, JODC makes no such assumptions and thus also makes it possible to compute a solution for non-smooth environments.

\paragraph{Rate of Convergence.} In the experiments, both algorithms converge to high-quality policy and environment pairs. However, we observe that DEPS converges faster in the MSD and in the microgrid environment. In the drone environment, the JODC algorithm initially performs better but it is quickly outperformed by the DEPS algorithm. The faster convergence of the DEPS algorithm can be explained by the way the gradients are computed in the two algorithms. For both methods, the computation of the gradients is based on the estimation of the expectation of a random variable using Monte-Carlo samples. In DEPS, the random variable is defined over the space of trajectories. On the other hand, in JODC, it is defined over the joint space of trajectories and environment parameters. For a given number of sampled trajectories, the update of the parameters in JODC is thus likely to be subject to more variance as the support of the random variable used for computing the gradients is much larger than the one used in DEPS. This is also the justification for our method being more sample efficient, as shown previously in numerical simulations.

\paragraph{Solution Quality.} In our experiments, DEPS outperformed JODC in all three environments. A first explanation may be that the gradient approximations in the JODC algorithm suffer from a high degree of variance, such that the gradient is unlikely to vanish when approaching an optimal solution. Second, we surmise that when approaching a (locally) optimal solution, the distribution over environments maintained by the JODC algorithm will shift its mass around parameters corresponding to this local optimum. The density probability function of environments in the neighborhood of an optimum, which is used when computing the gradients, then tends to take very large values, which may result in numerical issues. In practice, it seems to cause instabilities as we observe that the distribution over the environment parameters computed by JODC oscillates close to the (local) optimal solution computed with DEPS between iterations. This observation is directly reflected in the standard deviations of the final environment parameters learned with JODC, which are higher than those produced by the DEPS algorithm.


\section{Conclusions and Future Work}\label{sec:conclusions}
In this paper, we studied problems where an agent seeks to design and control discrete-time stochastic dynamical systems such that the resulting design and control policy jointly maximize the expected sum of rewards collected over a finite time horizon. In this context, the transition and reward functions are parametrized by environment parameters, while the policy is parametrized by policy parameters. In addition, these parametric functions are assumed known and differentiable with respect to their parameters. We introduced the DEPS algorithm, which is a deep RL algorithm combining policy gradient methods with model-based optimization techniques, to solve this problem. At its core, the algorithm iteratively updates the environment and policy parameters by approximating the gradient of the expected return with respect to these parameters via Monte-Carlo sampling and automatic differentiation and taking projected gradient ascent steps. The performance of the DEPS algorithm was empirically evaluated in three environments. The first environment is a mass-spring-damper system. The second and third environments, which are much more complex, are concerned with the design and control of an off-grid microgrid and a drone, respectively. In addition, for each environment, the performance of the DEPS algorithm is benchmarked against that of the so-called JODC algorithm \citep{schaff2019jointly}, which is a state-of-the-art algorithm designed to tackle joint design and control problems. Results show that the DEPS algorithm performs at least as well or better than the JODC algorithm in all three environments, consistently yielding solutions with higher returns in fewer iterations. In addition, in the first two environments, solutions produced by our algorithm are also compared with solutions produced by an algorithm that does not jointly optimize environment and policy parameters. The latter consists in designing a control policy \textit{a priori} and then optimizing environment parameters using derivative-free optimization methods so as to maximize the expected return. We highlighted that higher returns were achieved when joint optimization was performed. 

We identified three research directions that would be worth pursuing in future work. First, since the computational complexity of automatic differentiation is proportional to the length of the optimization horizon, the problem may become intractable for long horizons. A common solution applied in deep learning consists in truncating the back-propagation when computing gradients. Obtaining an analytical bound on the error when performing this approximation would be valuable for striking a trade-off between computational efficiency and solution quality. Second, the proposed method could also be combined with recent research in gradient-based direct policy search. In particular, the use of actor-critic methods, proximal policy optimization, etc., that are shown to result in stable learning and efficient exploration, could lead to better performance. Finally, in this paper we assume that we have direct access to the parametrized transition and reward functions of the system in addition to the disturbance distribution. These assumptions could be relaxed, and an approximate representation of these functions could be learned instead (e.g., using differentiable function approximators). This model could then be used in the DEPS algorithm for performing the usual parameter updates. A direct connection would then exist between the resulting algorithm and model-based RL techniques \citep{moerland2020model}, where an additional learning step is introduced in order to build an approximate model of the system from sampled trajectories. The latter class of methods has been successively applied on diverse problems \citep{serban2020bottleneck, bechtle2020curious, wu2020scalable}.

\section*{Acknowledgments}
The authors would like to thank Hatim Djelassi for valuable comments on this manuscript and François Cornet for his work on an early version of this article.  Adrien Bolland gratefully acknowledges the financial support of a research fellowship of the F.R.S.-FNRS. The authors also acknowledge the financial support of the Belgian government through the INTEGRATION project. Finally, the authors would like to thank the anonymous reviewers, whose comments and suggestions helped improve the clarity and quality of this manuscript.

\newpage
\appendix
\section{Analytical Derivation of the Gradient for Learning Optimal Environments} \label{ap:proof}

\paragraph{Theorem 1.} Let $(\St, \A, \Xi, P_0, f_\psi, \rho_\psi, P_\xi, T)$ and $\pi_\theta$ be an environment and a policy as defined in Section \ref{sec:problem_statement}. Additionally, let the functions $f_\psi$, $\rho_\psi$ and $P_\xi$ be continuously differentiable on the parameter space $\Psi$ and on the state space $\St$ for every action in $\A$ and every disturbance in $\Xi$. Furthermore, let the policy $\pi_\theta$ be continuously differentiable on its parameter space $\Theta$ and on the state space $\St$ for every action in $\A$ and for every time $t$. Let $V(\psi, \theta)$ be the expected cumulative reward of policy $\pi_\theta$, averaged over the initial states, for all $(\psi, \theta) \in \Psi\times\Theta$, as defined in equation \eqref{eq:exp_cum_rew_param}.
Then, the function $V$ exists, is bounded, and is continuously differentiable in the interior of $\Psi\times\Theta$.

\paragraph{Proof.} Let us first define the random variable associating the cumulative reward to a realization of a trajectory sampled from a policy in the environment for fixed parameter vectors $(\psi, \theta) \in \Psi\times\Theta$. We show that its expectation exists and is bounded for all $(\psi, \theta) \in \Psi\times\Theta$. Furthermore, $V(\psi, \theta)$ is defined by a parametric integral which we prove to be continuously differentiable for all $(\psi, \theta) \in \Psi\times\Theta$.

Let $\mathcal{R}_{\psi, \theta}$ be the real-valued random variable that, given $\psi\in \Psi$ and $\theta \in \Theta$, associates its cumulative reward to the realization of a trajectory. Thus, given a trajectory $\tau$, the random variable $\mathcal{R}_{\psi, \theta}$ takes value $R_{\psi, \theta}(\tau)$, as defined in equation \eqref{eq:cum_rew_traj}. Let $P_{\mathcal{R}_{\psi,\theta}}$ be the probability of this random variable. We can write:
\begin{align}
P_{\mathcal{R}_{\psi,\theta}}(\tau)
&= P_{\psi, \theta}(s_0, a_0, \xi_0, a_1, \xi_1, \dots, a_{T-1}, \xi_{T-1}) \\
&= P_0(s_0)\prod_{t=0}^{T-1} \pi_\theta(a_t|s_t, t) P_\xi(\xi_t| s_t, a_t) \; , \label{eq:probability_traj}
\end{align}
where the dependence upon $\psi$ is implicit through $s_{t+1} = f_\psi(s_t, a_t, \xi_t)$.
The expected cumulative reward given in equation \eqref{eq:exp_cum_rew_param} is the expectation of the random variable $\mathcal{R}_{\psi, \theta}$. If the expectation exists, it can therefore be written as:
\begin{multline}
V(\psi, \theta) = \int \big ( P_0(s_0)\prod_{t=0}^{T-1} \pi_\theta(a_t|s_t, t) P_\xi(\xi_t| s_t, a_t) \big ) \\ \big ( \sum_{t=0}^{T-1} \rho_\psi(s_t, a_t, \xi_t) \big )\: ds_0 da_0\dots da_{T-1} d\xi_0\dots d\xi_{T-1} \; , 
\end{multline}
or, more simply, as:
\begin{align}
V(\psi, \theta)
&= \int P_{\mathcal{R}_{\psi,\theta}}(\tau) R_{\psi, \theta}(\tau) d\tau \; . \label{eq:exp_cum_rew_int_param}
\end{align}

The theory of integration shows that a measurable function whose norm is upper-bounded almost-everywhere by that of an integrable function on a domain is itself integrable on this domain. Moreover, a random variable is measurable by definition and the reward function is bounded by $r_{max}$ such that: 
\begin{align}
\int |P_{\mathcal{R}_{\psi,\theta}}(\tau) R_{\psi, \theta}(\tau)| d\tau \leq \int P_{\mathcal{R}_{\psi,\theta}}(\tau)\: T\: r_{max} d\tau
\end{align}
In addition, by the Kolmogorov axioms, we know that:
\begin{align}
\int P_{\mathcal{R}_{\psi,\theta}}(\tau)\: T\: r_{max} d\tau = T\: r_{max} \; .
\end{align}
The integral defined by equation \eqref{eq:exp_cum_rew_int_param} thus exists and the function $V$ is bounded for all $(\psi, \theta) \in \Psi \times \Theta$ as follows:
\begin{align}
|\int P_{\mathcal{R}_{\psi,\theta}}(\tau) R_{\psi, \theta}(\tau) d\tau | \leq \int |P_{\mathcal{R}_{\psi,\theta}}(\tau) R_{\psi, \theta}(\tau)| d\tau \leq T\: r_{max} \; . \label{eq:exp_bound_cum_rew_int_param}
\end{align}

Finally, as a corollary to the Leibniz integral rule, a function defined as in equation \eqref{eq:exp_cum_rew_int_param} is continuously differentiable on the interior of the set $\Psi\times \Theta$ if $P_{\mathcal{R}_{\psi,\theta}} R_{\psi, \theta}(\tau)$ is continuously differentiable on the compact $\Psi\times\Theta$ for all trajectories $\tau \in X$, where $X = \St \times (\A \times \Xi)^T$ is the compact set of all trajectories. The latter is a composite function of $f_\psi$, $\rho_\psi$, $P_\xi$, and $\pi_\theta$. The composite function is thus continuously differentiable if the functions composing it are continuously differentiable, which is true by hypothesis. Furthermore, it implies that the partial derivative of the integral is equal to the integral of the partial derivative of the integrand.

\hfill$\square$

\paragraph{Corollary 1.} The function $V$, as defined in Theorem 1, exists, is bounded, and is continuously differentiable in the interior of $\Psi\times\Theta$ if $\A$ and/or $\Xi$ are discrete.

\paragraph{Proof.} Let us write the expression of the expectation \eqref{eq:exp_cum_rew_param} in the three cases depending on whether $\A$ and/or $\Xi$ are discrete and let us show that the different results of Theorem 1 are still valid.

\begin{enumerate}
	\item If $\A$ is discrete:
	\begin{multline}
V(\psi, \theta) = \int \sum_{(a_0, \dots a_{T-1}) \in \A^T} \big ( P_0(s_0)\prod_{t=0}^{T-1} \pi_\theta(a_t|s_t, t) P_\xi(\xi_t| s_t, a_t) \big ) \\ \big ( \sum_{t=0}^{T-1} \rho_\psi(s_t, a_t, \xi_t) \big )\: ds_0 d\xi_0\dots d\xi_{T-1}  \; . 
\end{multline}
	\item If $\Xi$ is discrete:
	\begin{multline}
V(\psi, \theta) = \int  \sum_{(\xi_0, \dots \xi_{T-1}) \in \Xi^T} \big ( P_0(s_0)\prod_{t=0}^{T-1} \pi_\theta(a_t|s_t, t) P_\xi(\xi_t| s_t, a_t) \big ) \\ \big ( \sum_{t=0}^{T-1} \rho_\psi(s_t, a_t, \xi_t) \big )\: ds_0 da_0\dots da_{T-1} \; .
\end{multline}
		\item If $\A$ and $\Xi$ are discrete:
	\begin{multline}
V(\psi, \theta) = \int \sum_{(a_0, \dots a_{T-1}) \in \A^T} \sum_{(\xi_0, \dots \xi_{T-1}) \in \Xi^T} \big ( P_0(s_0)\prod_{t=0}^{T-1} \pi_\theta(a_t|s_t, t) P_\xi(\xi_t| s_t, a_t) \big ) \\ \big ( \sum_{t=0}^{T-1} \rho_\psi(s_t, a_t, \xi_t) \big )\: ds_0 \; .
\end{multline}

\end{enumerate}
In the three cases, we can still bound the integral, as done in equation \eqref{eq:exp_bound_cum_rew_int_param}, and apply the corollary of the Leibniz integral rule if the integrand is continuously differentiable for all discrete values. Finally, by linearity of the differential operator, the operator can be distributed on the terms of the different sums when computing the derivative of the function $V$.

\hfill$\square$

\paragraph{Corollary 2.} The gradient of the function $V$ defined in equation \eqref{eq:exp_cum_rew_param} with respect to the parameter vector $\psi$ is such that:
\begin{multline}
\nabla_\psi  V(\psi, \theta)
= \E_{\underset{\xi_t \sim P_\xi(\cdot|s_t, a_t)}{\underset{a_t \sim \pi_\theta(\cdot|s_, t)}{s_0\sim P_0(\cdot)}}} \Big \{\Big (\sum_{t=0}^{T-1} \big( \nabla_s \log \pi_\theta(a_t|s, t)|_{s=s_t} + \nabla_s \log P_\xi(\xi_t| s, a_t)|_{s=s_t} \big ) \cdot \nabla_\psi s_t  \Big ) \\
\times \Big (\sum_{t=0}^{T-1} r_t \Big ) + \Big (\sum_{t=0}^{T-1} \nabla_\psi \rho_\psi(s, a_t, \xi_t)|_{s=s_t} + \nabla_s \rho_\psi(s, a_t, \xi_t)|_{s=s_t} \cdot \nabla_\psi s_t \Big ) \Big \} \label{eq:apx_grad_v_psi}  ,
\end{multline}
where:
\begin{align}
\nabla_\psi s_t &= (\nabla_s f_\psi)(s, a_{t-1}, \xi_{t-1})|_{s=s_{t-1}} \cdot \nabla_\psi s_{t-1} + (\nabla_\psi f_\psi)(s, a_{t-1}, \xi_{t-1})|_{s=s_{t-1}} \label{eq:apx_grad_s}  \; ,
\end{align}
with $\nabla_\psi s_0 = 0$.

\paragraph{Proof.} 
To compute this gradient, we first apply the product rule for gradients to equation \eqref{eq:exp_cum_rew_param}. Afterwards, we exploit the identity $\nabla f = f \nabla \log f$ (which we refer to as the \textit{log-derivative trick}) that holds if $f$ is a continuously differentiable function.
\begin{align}
\nabla_\psi  V(\psi, \theta)
&= \int (\nabla_\psi P_{\mathcal{R}_{\psi,\theta}}(\tau)) R_{\psi, \theta}(\tau) d\tau + \int P_{\mathcal{R}_{\psi,\theta}}(\tau) (\nabla_\psi R_{\psi, \theta}(\tau)) d\tau \\
&= \int P_{\mathcal{R}_{\psi,\theta}}(\tau) (\nabla_\psi \log P_{\mathcal{R}_{\psi,\theta}}(\tau)) R_{\psi, \theta}(\tau) d\tau + \int P_{\mathcal{R}_{\psi,\theta}}(\tau) (\nabla_\psi R_{\psi, \theta}(\tau)) d\tau\\
&= \E_{\underset{\xi_t \sim P_\xi(\cdot|s_t, a_t)}{\underset{a_t \sim \pi_\theta(\cdot|s_t, t)}{s_0\sim P_0(\cdot)}}} \{(\nabla_\psi \log P_{\mathcal{R}_{\psi,\theta}}(\tau)) R_{\psi, \theta}(\tau) + (\nabla_\psi R_{\psi, \theta}(\tau)) \} \; . \label{eq:apx_exp_gradient_psi}
\end{align}
By applying the logarithmic operator to both sides of equation \eqref{eq:probability_traj}, we have:
\begin{align}
\log P_{\mathcal{R}_{\psi,\theta}}(\tau) &= \log P_0(s_0) + \sum_{t=0}^{T-1} \log \pi_\theta(a_t|s_t, t) + \sum_{t=0}^{T-1} \log P_\xi(\xi_t| s_t, a_t) \; . \label{eq:apx_log_probability_traj} 
\end{align}

Let $\cdot$ denote the dot product operator. Using the chain rule formula together with equation \eqref{eq:cum_rew_traj}, we can write:
\begin{align}
\nabla_\psi \log \pi_\theta(a_t|s_t, t) &=  \nabla_s \log \pi_\theta(a_t|s, t)|_{s=s_t} \cdot \nabla_\psi s_t \\
\nabla_\psi \log P_\xi(\xi_t| s_t, a_t) &= \nabla_s \log P_\xi(\xi_t| s, a_t)|_{s=s_t} \cdot \nabla_\psi s_t \\
\nabla_\psi \rho_\psi(s_t, a_t, \xi_t) &= \nabla_\psi \rho_\psi(s, a_t, \xi_t)|_{s=s_t} + \nabla_s \rho_\psi(s, a_t, \xi_t)|_{s=s_t} \cdot \nabla_\psi s_t \; ,
\end{align}		
where:
\begin{align}
\nabla_\psi s_t &= (\nabla_s f_\psi)(s, a_{t-1}, \xi_{t-1})|_{s=s_{t-1}} \cdot \nabla_\psi s_{t-1} + (\nabla_\psi f_\psi)(s, a_{t-1}, \xi_{t-1})|_{s=s_{t-1}} \; ,
\end{align}
with $\nabla_\psi s_0 = 0$.

Finally, combining the previous results with equations \eqref{eq:apx_exp_gradient_psi} and \eqref{eq:apx_log_probability_traj}, we have:
\begin{multline}
\nabla_\psi  V(\psi, \theta)
= \E_{\underset{\xi_t \sim P_\xi(\cdot|s_t, a_t)}{\underset{a_t \sim \pi_\theta(\cdot|s_, t)}{s_0\sim P_0(\cdot)}}} \Big \{\Big (\sum_{t=0}^{T-1} \big( \nabla_s \log \pi_\theta(a_t|s, t)|_{s=s_t} + \nabla_s \log P_\xi(\xi_t| s, a_t)|_{s=s_t} \big ) \cdot \nabla_\psi s_t  \Big ) \\
\times \Big (\sum_{t=0}^{T-1} r_t \Big ) + \Big (\sum_{t=0}^{T-1} \nabla_\psi \rho_\psi(s_t, a_t, \xi_t) \Big) \Big \} \; .
\end{multline}

\hfill$\square$

\paragraph{Corollary 3.} The gradient of the function $V$, defined in equation \eqref{eq:exp_cum_rew_param}, with respect to the parameter vector $\theta$ is given by:
\begin{align}
\nabla_\theta  V(\psi, \theta)
&=  \E_{\underset{\xi_t \sim P_\xi(\cdot|s_t, a_t)}{\underset{a_t \sim \pi_\theta(\cdot|s_t, t)}{s_0\sim P_0(\cdot)}}} \{ (\sum_{t=0}^{T-1} \nabla_\theta \log \pi_\theta(a_t|s_t, t)) (\sum_{t=0}^{T-1} r_t) \} \;  .
\end{align}

\paragraph{Proof.}
Using techniques similar to the ones used in the proof of Corollary 1 for the gradient with respect to $\theta$, we find:
\begin{align}
\nabla_\theta  V(\psi, \theta)
&= \E_{\underset{\xi_t \sim P_\xi(\cdot|s_t, a_t)}{\underset{a_t \sim \pi_\theta(\cdot|s_t, t)}{s_0\sim P_0(\cdot)}}} \{(\nabla_\theta \log P_{\mathcal{R}_{\psi,\theta}}(\tau)) R_{\psi, \theta}(\tau) \} \\
&=  \E_{\underset{\xi_t \sim P_\xi(\cdot|s_t, a_t)}{\underset{a_t \sim \pi_\theta(\cdot|s_t, t)}{s_0\sim P_0(\cdot)}}} \{ (\sum_{t=0}^{T-1} \nabla_\theta \log \pi_\theta(a_t|s_t, t)) (\sum_{t=0}^{T-1} r_t) \} \label{eq:apx_grad_v_theta} \; .
\end{align}

\hfill$\square$

\paragraph{Theorem 2.} Let $(\St, \A, \Xi, P_0, f_\psi, \rho_\psi, P_\xi, T)$ and $\pi_\theta$ be an environment and a policy, respectively, as defined in Section \ref{sec:problem_statement}. Let $V(\psi, \theta)$ be the expected cumulative reward of policy $\pi_\theta$ averaged over the initial states, as defined in equation \eqref{eq:exp_cum_rew_param}. Let $\mathcal{D} = \{h^m| m = 0, \dots, M-1\}$ be a set of $M$ histories sampled independently and identically from the policy $\pi_\theta$ in the environment. Let $\mathcal{L}$ be a loss function such that,  $\forall (\psi, \theta) \in \Psi\times\Theta$:
\begin{multline}
\mathcal{L}(\psi, \theta) = -\frac{1}{M} \sum_{m=0}^{M-1} \Big (\sum_{t=0}^{T-1} \log \pi_\theta(a_t^m|s_t^m, t) + \log P_\xi(\xi_t^m| s_t^m, a_t^m) \big ) 
\\ \times \big ((\sum_{t=0}^{T-1} r_t^m) - B \big ) 
+ \big (\sum_{t=0}^{T-1} \rho_\psi(s_t^m, a_t^m, \xi_t^m) \big) \Big) \; ,
\end{multline}
where $B$ is a constant value called the baseline.
The gradients with respect to $\psi$ and $\theta$ of the loss function are unbiased estimators of the gradients of the function $V$ as defined in equation \eqref{eq:exp_cum_rew_param} with opposite directions, i.e., they are such that: 
\begin{align}
\E_{\underset{\xi_t \sim P_\xi(\cdot|s_t, a_t)}{\underset{a_t \sim \pi_\theta(\cdot|s_t, t)}{s_0\sim P_0(\cdot)}}} \{ \nabla_\psi \mathcal{L}(\psi, \theta)\} &= - \nabla_\psi  V(\psi, \theta) \\
\E_{\underset{\xi_t \sim P_\xi(\cdot|s_t, a_t)}{\underset{a_t \sim \pi_\theta(\cdot|s_t, t)}{s_0\sim P_0(\cdot)}}} \{ \nabla_\theta \mathcal{L}(\psi, \theta) \} &= - \nabla_\theta  V(\psi, \theta)  \; .
\end{align}

\paragraph{Proof.} Let us first rewrite the loss function using the notations of Theorem 1. We have:
\begin{align}
\mathcal{L}(\psi, \theta) = -\frac{1}{M} \sum_{m=0}^{M-1} (\log P_{\mathcal{R}_{\psi,\theta}}(\tau^m) - \log P_0(s_0^m)) \big ((\sum_{t=0}^{T-1} r_t^m) - B \big )  + (R_{\psi, \theta}(\tau^m)) \; .
\end{align}
The expectation of the gradient with respect to $\psi$ is given by:
\begin{multline}
\E_{\underset{\xi_t \sim P_\xi(\cdot|s_t, a_t)}{\underset{a_t \sim \pi_\theta(\cdot|s_t, t)}{s_0\sim P_0(\cdot)}}} \{ \nabla_\psi \mathcal{L}(\psi, \theta) \} = \E_{\underset{\xi_t \sim P_\xi(\cdot|s_t, a_t)}{\underset{a_t \sim \pi_\theta(\cdot|s_t, t)}{s_0\sim P_0(\cdot)}}} \{ -\frac{1}{M} \sum_{m=0}^{M-1} \nabla_\psi(\log P_{\mathcal{R}_{\psi,\theta}}(\tau^m) - \log P_0(s_0^m)) \\
\times \big ((\sum_{t=0}^{T-1} r_t^m) - B \big )  + \nabla_\psi(R_{\psi, \theta}(\tau^m)) \} \; .
\end{multline}
Observing that every term in the sum has the same expectation and that $\nabla_\psi \log P_0(s_0^m) = 0$, we can write:
\begin{align}
\E_{\underset{\xi_t \sim P_\xi(\cdot|s_t, a_t)}{\underset{a_t \sim \pi_\theta(\cdot|s_t, t)}{s_0\sim P_0(\cdot)}}} \{ \nabla_\psi \mathcal{L}(\psi, \theta) \} = -\E_{\underset{\xi_t \sim P_\xi(\cdot|s_t, a_t)}{\underset{a_t \sim \pi_\theta(\cdot|s_t, t)}{s_0\sim P_0(\cdot)}}} \{ \nabla_\psi(\log P_{\mathcal{R}_{\psi,\theta}}(\tau))
\times \big ((\sum_{t=0}^{T-1} r_t) - B \big )  + \nabla_\psi(R_{\psi, \theta}(\tau)) \}  \; .
\end{align}
Moreover, using the log-derivative trick:
\begin{align}
\E_{\underset{\xi_t \sim P_\xi(\cdot|s_t, a_t)}{\underset{a_t \sim \pi_\theta(\cdot|s_t, t)}{s_0\sim P_0(\cdot)}}} \{ \nabla_\psi(\log P_{\mathcal{R}_{\psi,\theta}}(\tau)) B \} 
= \nabla_\psi \E_{\underset{\xi_t \sim P_\xi(\cdot|s_t, a_t)}{\underset{a_t \sim \pi_\theta(\cdot|s_t, t)}{s_0\sim P_0(\cdot)}}} \{B \} = 0 \; ,
\end{align}
such that:
\begin{align}
\E_{\underset{\xi_t \sim P_\xi(\cdot|s_t, a_t)}{\underset{a_t \sim \pi_\theta(\cdot|s_t, t)}{s_0\sim P_0(\cdot)}}} \{ \nabla_\psi \mathcal{L}(\psi, \theta)\} &= - \nabla_\psi  V(\psi, \theta) \; .
\end{align}

Equivalently, the expectation of the gradient with respect to $\theta$ is given by:
\begin{multline}
\E_{\underset{\xi_t \sim P_\xi(\cdot|s_t, a_t)}{\underset{a_t \sim \pi_\theta(\cdot|s_t, t)}{s_0\sim P_0(\cdot)}}} \{ \nabla_\theta \mathcal{L}(\psi, \theta) \} = -\E_{\underset{\xi_t \sim P_\xi(\cdot|s_t, a_t)}{\underset{a_t \sim \pi_\theta(\cdot|s_t, t)}{s_0\sim P_0(\cdot)}}} \{ \nabla_\theta(\log P_{\mathcal{R}_{\psi,\theta}}(\tau)) \\
\times \big ((\sum_{t=0}^{T-1} r_t) - B \big )  + \nabla_\theta(R_{\psi, \theta}(\tau)) \}  \; .
\end{multline}
The expectation of the term relative to the baseline is zero:
\begin{align}
\E_{\underset{\xi_t \sim P_\xi(\cdot|s_t, a_t)}{\underset{a_t \sim \pi_\theta(\cdot|s_t, t)}{s_0\sim P_0(\cdot)}}} \{ \nabla_\theta(\log P_{\mathcal{R}_{\psi,\theta}}(\tau)) B \} 
= \nabla_\theta \E_{\underset{\xi_t \sim P_\xi(\cdot|s_t, a_t)}{\underset{a_t \sim \pi_\theta(\cdot|s_t, t)}{s_0\sim P_0(\cdot)}}} \{B \} = 0 \; .
\end{align}
Furthermore, the gradient of the reward function with respect to $\theta$ is zero:
\begin{align}
\E_{\underset{\xi_t \sim P_\xi(\cdot|s_t, a_t)}{\underset{a_t \sim \pi_\theta(\cdot|s_t, t)}{s_0\sim P_0(\cdot)}}} \{ \nabla_\theta(R_{\psi, \theta}(\tau)) \} = \E_{\underset{\xi_t \sim P_\xi(\cdot|s_t, a_t)}{\underset{a_t \sim \pi_\theta(\cdot|s_t, t)}{s_0\sim P_0(\cdot)}}} \{  (\sum_{t=0}^{T-1} \nabla_\theta \rho_\psi(s_t, a_t, \xi_t) \} = 0 \; .
\end{align}
We thus have that:
\begin{align}
\E_{\underset{\xi_t \sim P_\xi(\cdot|s_t, a_t)}{\underset{a_t \sim \pi_\theta(\cdot|s_t, t)}{s_0\sim P_0(\cdot)}}} \{ \nabla_\theta \mathcal{L}(\psi, \theta) \} 
&= -\E_{\underset{\xi_t \sim P_\xi(\cdot|s_t, a_t)}{\underset{a_t \sim \pi_\theta(\cdot|s_t, t)}{s_0\sim P_0(\cdot)}}} \{ \nabla_\theta(\log P_{\mathcal{R}_{\psi,\theta}}(\tau))
\times \big (\sum_{t=0}^{T-1} r_t \big ) \} \\
&= -\E_{\underset{\xi_t \sim P_\xi(\cdot|s_t, a_t)}{\underset{a_t \sim \pi_\theta(\cdot|s_t, t)}{s_0\sim P_0(\cdot)}}} \{ \nabla_\theta(\sum_{t=0}^{T-1} \log \pi_\theta(a_t| s_t, t))
\times \big (\sum_{t=0}^{T-1} r_t \big ) \}  \; ,
\end{align}
and thus:
\begin{align}
\E_{\underset{\xi_t \sim P_\xi(\cdot|s_t, a_t)}{\underset{a_t \sim \pi_\theta(\cdot|s_t, t)}{s_0\sim P_0(\cdot)}}} \{ \nabla_\theta \mathcal{L}(\psi, \theta) \} = - \nabla_\theta  V(\psi, \theta) \; .
\end{align}

\hfill$\square$

\paragraph{Corollary 4.} The gradient of the loss function, defined in equation \eqref{eq:loss_function}, with respect to $\theta$ corresponds to the opposite of the update direction computed with the REINFORCE algorithm \citep{williams1992simple} averaged over $M$ simulations.

\paragraph{Proof.} The gradient of the loss function with respect to $\theta$ is given by:
\begin{align}
\nabla_\theta \mathcal{L}(\psi, \theta) = - \sum_{m=0}^{M-1} \Big ( \nabla_\theta(\log P_{\mathcal{R}_{\psi,\theta}}(\tau^m)) \times \big (R_{\psi,\theta}(\tau^m) - B \big ) \Big ) \; .
\end{align}

The gradient is the opposite of the average over $M$ trajectories of the update direction of the REINFORCE algorithm \citep{williams1992simple}.

\hfill$\square$

\newpage
\section{The Direct Environment and Policy Search Algorithm} \label{ap:algos}

\begin{algorithm}[H]
	\caption{DEPS}
	\label{algo:system_optimization}
	\begin{algorithmic}
		\State \textbf{function} \texttt{Optimize}($(\St, \A, \Xi, P_0, f_\psi, \rho_\psi, P_\xi, T)$, $\pi_\theta$, $\Pi_{\Psi}$, $\Pi_{\Theta}$)
		\State \textbf{Parameter} Number of gradient ascent steps $N$
		\State \textbf{Parameter} Batch size $M$
		\State \textbf{Parameter} Learning rate $\alpha$
		\ForAll {$n \in \{0, \dots, N-1\}$}
		\ForAll {$m \in \{0, \dots, M-1\}$}
		\State $h =$ \texttt{GenerateHistory}($(\St, \A, \Xi, P_0, f_\psi, \rho_\psi, P_\xi, T)$, $\pi_\theta$)
		\State Add $h$ to the set $\mathcal{D}$
		\EndFor 	
		\State Compute the baseline using the histories $B = \frac{1}{m} \sum_{m=0}^{M-1} \sum_{t=0}^{T-1} r_t$
		\State Differentiate equation \eqref{eq:loss_function} for estimating the gradients equations \eqref{eq:grad_v_psi} and \eqref{eq:grad_v_theta} using $\mathcal{D}$
		\State $(\psi, \theta) =$ \texttt{VanillaGradientAscent}($\psi$, $\theta$, $\alpha$, $\hat{\nabla}_\psi V(\psi, \theta)$, $\hat{\nabla}_\theta V(\psi, \theta)$)
		\State $\psi \leftarrow \Pi_{\Psi} (\psi)$
		\State $\theta \leftarrow \Pi_{\Theta}(\theta)$
		\EndFor 
		\State
		\Return $(\psi, \theta)$
		\State
		\State \textbf{function} \texttt{GenerateHistory}($(\St, \A, \Xi, P_0, f_\psi, \rho_\psi, P_\xi, T)$, $\pi_\theta$)
		\State Sample an initial state: $s_0 \sim P_0(\cdot)$
		\ForAll {$t \in \{0, \dots, T-1\}$}
		\State $a_t \sim \pi_\theta(\cdot |s_t, t)$
		\State $\xi_t \sim P_\xi(\cdot |s_t, a_t)$
		\State $s_{t+1} = f_\psi(s_t, a_t, \xi_t)$
		\State $r_t = \rho_\psi(s_t, a_t, \xi_t)$
		\EndFor
		\State $h = (s_0, a_0, \xi_0, r_0, a_1, \xi_1, \dots, a_{T-1}, \xi_{T-1}, r_{T-1}) $
		\State
		\Return $h$
		\State
		\State \textbf{function} \texttt{VanillaGradientAscent}($\psi$, $\theta$, $\alpha$, $\hat{\nabla}_\psi V(\psi, \theta)$, $\hat{\nabla}_\theta V(\psi, \theta)$)
		\State $\psi \leftarrow \psi + \alpha \cdot \hat{\nabla}_\psi V(\psi, \theta)$
		\State $\theta \leftarrow \theta + \alpha \cdot \hat{\nabla}_\theta V(\psi, \theta)$
		\State
		\Return $(\psi, \theta)$
	\end{algorithmic}
\end{algorithm}

\newpage
\section{Jointly Optimizing the Design and the Policy of a System} \label{ap:joint_opt}
In this section, we review two alternatives to the DEPS algorithm. First, we discuss the Joint Optimization of Design and Control (JODC) algorithm \citep{schaff2019jointly}, an algorithm that jointly optimizes a control policy and a distribution over the environment parameters in order to maximize the return of the policy on expectation over the distribution of environments. Then, we discuss a method that consists in designing a control policy \textit{a priori} and then optimizing environment parameters using derivative-free optimization methods.

\subsection{Joint Optimization of Design and Control Algorithm}
The JODC algorithm is an RL algorithm for jointly optimizing the design and the control policy of an environment \citep{schaff2019jointly}. The formalization is nevertheless slightly different from our problem statement equation \eqref{eq:problem_statement}. In contrast to the DEPS algorithm, the JODC algorithm does not optimize the environment parameters directly but instead optimizes the parameters of a distribution defined over these environment parameters in order to maximize the expected return of the policy on expectation over the distribution of environments. 

Formally, let $p_\phi$ be a distribution defined over the set of environment parameters $\Psi$. The distribution $p_\phi$ is parametrized by the vector $\phi \in \Phi \subsetneq \R^{d_\Phi}$, where $d_\Phi$ is the dimension of the space of parameters of such distributions. In the following, $p_\phi$ is referred to as the environment policy. Furthermore, let us use the same notations as in Section \ref{sec:problem_statement}. Thus, $\pi_\theta$ is a parametrized control policy and $V(\psi, \theta)$ is the expected return of this policy in the environment parametrized by $\psi$. The pair $(\phi^*, \theta^*)$ is optimal if it is such that:
\begin{align}
\phi^*, \theta^* &\in \underset{\phi \in \Phi, \theta \in \Theta}{\mathrm{argmax\;}} \E_{\psi \sim p_\phi(\cdot)} \{ V(\psi, \theta) \} \; .
\end{align}
Using the log-derivative trick, the gradient with respect to $\phi$ can be computed as follows:
\begin{align}
\nabla_\phi  \E_{\psi \sim p_\phi(\cdot)} \{ V(\psi, \theta) \}
&=  \E_{\underset{\underset{\xi_t \sim P_\xi(\cdot|s_t, a_t)}{\underset{a_t \sim \pi_\theta(\cdot|s_t, t)}{s_0\sim P_0(\cdot)}}}{\psi \sim p_\phi(\cdot)}} \{ (\nabla_{\phi} \log p_\phi(\psi)) (\sum_{t=0}^{T-1} r_t) \} \;  . \label{eq:jodc_grad_phi}
\end{align}
The JODC algorithm then approximates this gradient via Monte-Carlo sampling and exploits it to update the environment policy parameters. More precisely, a batch of $M$ environments $\mathcal{E} = \{ \psi^m | m=0,\dots, M-1 \}$ is first sampled from $p_\phi$. Then, the control policy is executed for each environment in the batch that was sampled. Afterwards, the parameters of the environment policy are updated by stochastic gradient ascent using the sampled trajectories to approximate the expectation in equation \eqref{eq:jodc_grad_phi}. In practice, the Adam algorithm \citep{kingma2014adam} is used and gradients are computed by automatic differentiation of the following loss function:
\begin{align}
	\mathcal{L} = - \frac{1}{M} \sum_{m=0}^{M-1} \log ( p_\phi(\psi^m) ) \big ( \sum_{t=0}^{T-1} r_t \big ) \; .
\end{align}
Finally, the control policy parameters are updated by applying the proximal policy optimization (PPO) algorithm \citep{schulman2017proximal} using the sampled trajectories. This algorithm is a variant of the vanilla stochastic gradient ascent of the REINFORCE algorithm that has been shown to perform well for learning control policies in complex environments. The main steps of the JODC algorithm are summarized in Algorithm \ref{algo:jodc_algo}.
\begin{algorithm}[H]
	\caption{JODC}
	\label{algo:jodc_algo}
	\begin{algorithmic}
		\Loop
		\State Sample $M$ environment parameters $\phi^m$ from $p_\phi$
		\State Execute the control policy $\pi_\theta$ in each sampled environment	
		\State Autodifferentiate $\mathcal{L}$ and apply Adam for updating $\phi$
		\State Apply PPO for updating $\theta$
		\EndLoop
	\end{algorithmic}
\end{algorithm}

Let us note that \cite{schaff2019jointly} applied two additional changes compared to the pseudocode displayed in Algorithm \ref{algo:jodc_algo}. First, they pre-fit the policy based on trajectories generated in environments sampled at random. Second, they make the policy explicitly dependent on the environment parameters $\psi$ as well. 

\subsection{System Optimization with Rule-Based Policies}
The second method is a decomposition of the design and the policy optimization processes as follows. First, we design a policy \textit{a priori} based on expert knowledge. For example, the policy could be a rule-based policy or a model-based controller (e.g., a linear-quadratic regulator or a model predictive controller). The environments are then optimized in order to find the parameters maximizing the reward given the policy. Formally, let $\pi(a_t|s_t, t; \psi)$ be the policy that gives the probability of selecting action $a_t$ when the system parametrized by $\psi$ is in state $s_t$ at time $t$. The optimal environment parameters are then computed by solving the following optimization problem:
\begin{align}
\psi^* &\in \underset{\psi \in \Psi}{\mathrm{argmax\;}} \E_{\underset{\xi_t \sim P_\xi(\cdot|s_t, a_t)}{\underset{a_t \sim \pi(\cdot|s_t, t; \psi)}{s_0\sim P_0(\cdot)}}} \{ \sum_{t=0}^{T-1} r_t \} \\
s_{t+1} &= f_\psi(s_t, a_t, \xi_t) \\
r_t &= \rho_\psi(s_t, a_t, \xi_t) \; .
\end{align}
In practice, this objective is pursued using a derivative-free optimization method and computing a Monte-Carlo estimation of the objective function at each point estimation of $\psi$.


\newpage
\section{Mass-Spring-Damper Environment} \label{ap:msd}

\begin{figure}[H]
	\centering
	\includegraphics[width=0.5\textwidth]{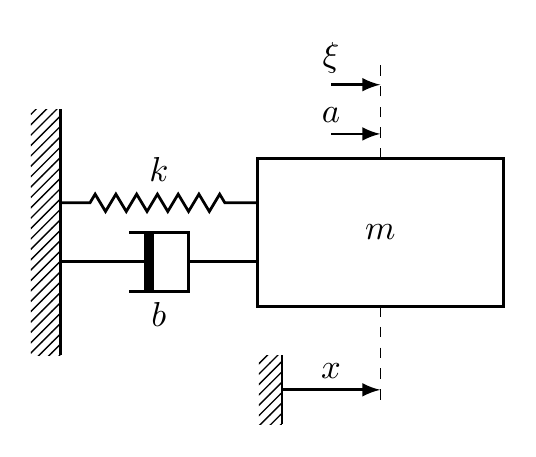}
	\caption{Mass-Spring-Damper system.}
	\label{fig:msd_sysem}
\end{figure}

Let us consider a Mass-Spring-Damper (MSD) system where an object, modeled as a point mass, is attached to a spring and a damper, as represented in Figure \ref{fig:msd_sysem}. In addition to the forces applied by the spring and the damper, the object is subject to a random force that depends on its position and velocity. An agent interacts with the system by applying a control force on the object. The control objective of the agent is to stabilize the spring at a fixed reference position. The agent also seeks to optimize a design function parametrized by the Hooke constant of the spring and the damping coefficient of the damper. Hence, the objective of the joint design and control problem is to optimize these objectives simultaneously.

\paragraph{Optimization Horizon.} The optimization horizon is denoted by $T=100$ and is equal to the number of actions to be taken throughout the discrete decision process.

\paragraph{State Space.} The state $s_t$ at time $t$ is described by two variables, namely the position $x_t$ and the velocity $v_t$ of the object. The position is positive when the spring is extended (i.e., elongated to the right in Figure \ref{fig:msd_sysem}) and negative when it is compressed (i.e., elongated to the left in Figure \ref{fig:msd_sysem}). In addition, the velocity is positive when the mass moves to the right in Figure \ref{fig:msd_sysem} and negative when it moves to the left. The state space of the system is thus the following:
\begin{align}
\St = \R^2 \; .
\end{align}

\paragraph{Initial State Distribution.} The initial state $s_0 = (x_0, v_0)$ is drawn uniformly at random from a subset of the state space $\St_0 = [x_{0, min}, x_{0, max}]\times[v_{0, min}, v_{0, max}] \subsetneq \St$.

\paragraph{Action Space.} At time $t$, an agent can apply a control force on the object. In the most general setting, this force may take values in $\mathbb{R}$. However, in this paper, we will consider that the agent can only apply a force selected from the following discrete set:
\begin{align}
\A = \{-0.3, -0.1, 0.0, 0.1, 0.3\}\; .
\end{align}
\paragraph{Disturbance Space.} We consider a stochastic version of the problem where a real-valued disturbance $\xi_t$ is added to the action $a_t$ at each time step $t$, such that the input effectively applied to the system is $a_t + \xi_t$. In this context, we have:
\begin{align}
\xi_t \in \Xi &= \R \; . 
\end{align}

\paragraph{Disturbance Distribution.} The disturbance at time $t$ is sampled from a normal distribution centered at the current position $x_t$, and whose standard deviation is a linear combination of the magnitudes of the action $a_t$ and the velocity $v_t$:
\begin{align}
P_\xi(\xi_t|s_t, a_t) &= \mathcal{N}(\xi_t \big |x_t, 0.1 \times |a_t| + |v_t| + \epsilon) \; , 
\end{align}
where $\epsilon$ is a constant equal to $10^{-6}$ ensuring that the standard deviation of the distribution remains strictly positive. Note that the disturbance is endogenous in the sense that it depends on the system state. This also introduces a feedback loop that can render the system unstable.

\paragraph{Transition Function.} Let $x$ denote the position of the object. The spring has Hooke constant $k$ and the damper has damping coefficient $b$. The force exerted by the spring on the object is in the direction opposite to its elongation and the damping force acts in the direction opposite to its motion. Furthermore, the system is subject to a control force as well as a random perturbation. Both are proportional to the mass $m$, and are obtained by multiplying the action $a$ and the disturbance $\xi$ by $m$, respectively. Then, the continuous-time dynamics of the system are described by Newton's second law as:
\begin{align}
m \ddot x = - k x - b \dot x + m a + m \xi \; ,
\end{align}
which can equivalently be written as:
\begin{align}
\ddot x + 2 \zeta \omega \dot x + \omega^2 x = a + \xi \; ,  \label{eq:msd_ode}
\end{align}
where:
\begin{align}
\omega &= \sqrt{\frac{k}{m}}\\
\zeta &= \frac{b}{2m\omega} \; .
\end{align}

In order to obtain the transition function of the discrete-time system, the optimization horizon is split into a set of successive time periods. Each time period $t$ has a duration of $\Delta t = 50$ms. Then, the transition dynamics are obtained from the analytical solution of equation \eqref{eq:msd_ode}. More precisely, let $g$ be the function computing the new position $x_{t+1}$ of the point mass based on $x_t$ and $v_t$ after a period $\tau = \Delta t$ during which the constant inputs $a_t$ and $\xi_t$ are applied. The transition function $f$ thus updates the state variables as follows: 
\begin{align}
x_{t+1} &= g(x_t, v_t, a_t, \xi_t, \Delta t) \\
v_{t+1} &= \frac{\partial g}{\partial \tau}(x_t, v_t, a_t, \xi_t, \tau)|_{\tau=\Delta t} \; , 
\end{align}
where:
\begin{multline}
g(x_t, v_t, a_t, \xi_t, \tau) = \frac{a_t + \xi_t}{\omega^2} + \exp(-\zeta \omega \tau) \times \\
\left \{
\begin{array}{r l}
 (x_t - \frac{a_t + \xi_t}{\omega^2}) \cosh(\sqrt{\zeta^2 - 1}\omega \tau) + \frac{\frac{v_t}{\omega} + \zeta (x_t - \frac{a_t + \xi_t}{\omega^2})}{\sqrt{\zeta^2 - 1}} \sinh(\sqrt{\zeta^2 - 1}\omega \tau) \; , &\text{if } \zeta > 1 \\
(x_t - \frac{a_t + \xi_t}{\omega^2}) + \big( v_t + \omega (x_t - \frac{a_t + \xi_t}{\omega^2}) \big )\tau \; ,   &\text{if } \zeta = 1 \\
 (x_t - \frac{a_t + \xi_t}{\omega^2}) \cos(\sqrt{1 - \zeta^2}\omega \tau) + \frac{\frac{v_t}{\omega} + \zeta (x_t - \frac{a_t + \xi_t}{\omega^2})}{\sqrt{1 - \zeta^2}} \sin(\sqrt{1 - \zeta^2}\omega \tau) \; , &\text{if } 0 < \zeta < 1 \; .
\end{array}
\right . 
\end{multline}

\paragraph{Reward Function.} The reward function is defined as:
\begin{align}
\rho(a_t, s_t, \xi_t) &= \exp \Big(-|x_t - x_{ref}| - (\omega - c_\omega)^2 - (\zeta - c_\zeta)^2 - \prod_{k=1}^K (\phi_k - c_k)^2\Big) \; , 
\end{align}
where $\omega$, $\zeta$ and $\phi_k$, $k = 1,\ldots, K$, are parameters of the system that need to be optimized. Furthermore $x_{ref}$, $c_\omega$, $c_\zeta$, $K$ and $c_k$ are constant values. Note that the reward function does not depend on the disturbance.

The first term in the exponential will be minimized if the mass is stabilized at some reference position $x_{ref}$. The second and third terms are minimized if the parameters $\omega$ and $\zeta$ are equal to $c_\omega$ and $c_\zeta$, respectively. The last term is a strictly positive function that is minimized if at least one of the parameters $\phi_k$ is equal to the corresponding constant $c_k$. This term is a non-convex function introduced to artificially increase the size of the parameter space and the complexity of the reward function. Simultaneously minimizing these terms is equivalent to maximizing the reward. Furthermore, since the reward function is the exponential of an expression only taking negative values, the reward is upper-bounded by $r_{max} = 1$. Hence, the expected return is upper-bounded by $T$.

\paragraph{Parametrized MSD Environment.} A parametrized MSD environment is an 8-tuple $(\St, \A, \Xi, P_0, f_\psi, \rho_\psi, P_\xi, T)$ parametrized by the real vector $\psi = (\omega, \zeta, \phi_0, \phi_1, \phi_2) \in \R^{+2} \times \R^3$.

\paragraph{Rule-Based Policies.} Several rule-based policies can be envisaged for the MSD system. First, let $a_{eq}$ be the steady-state normalized force that must be applied on the system in order to maintain the mass at the reference position $x_{ref}$, assuming that the disturbance is equal to zero. This normalized force is given by: 
\begin{align}
	a_{eq} &= \omega^2 x_{ref} \; .
\end{align}
The first rule-based policy $\pi_{msd, 1}$ makes use of a categorical distribution over the actions $a_t \in \A$ with expectation $a_{eq}$ and the smallest possible standard deviation. Let $a_{lower}$ and $a_{upper}$ be such that:
\begin{align}
	a_{lower} &= \underset{a \in \A}{\mathrm{argmax}} \{ a | a < a_{eq} \} \\
	a_{upper} &= \underset{a \in \A}{\mathrm{argmin}} \{ a | a \geq a_{eq} \} \; .
\end{align}
Then, to any state $s_t$, the first rule-based (stochastic) policy $\pi_{msd, 1}$ associates an action $a_t \in \{a_{lower}, a_{upper}\}$ by sampling from the following distribution:
\begin{align}
\left \{
\begin{array}{r c l}
 a_{lower} \; , &\text{ w.p. }& \frac{a_{upper} - a_{eq}}{a_{upper} - a_{lower}} \\
 a_{upper} \; , &\text{ w.p. }& \frac{ a_{eq} - a_{lower}}{a_{upper} - a_{lower}} \; .
\end{array}
\right . 
\end{align}
For the second rule-based policy, we propose a policy that seeks to move the point mass towards the reference position $x_{ref}$ if it happens to be outside of a neighborhood of $x_{ref}$. Assuming $x_{ref}$ is positive and takes the value provided in Table~\ref{tab:msd_input_param}, let $x_{push}$ and $x_{pull}$ be the bounds of the interval defining the neighborhood of  $x_{ref}$, such that $x_{push} \leq x_{ref} \leq x_{pull}$. Let $a_{push}$ and $a_{pull}$ be the actions taken when the position of the mass $x_t$ is smaller than $x_{push}$ and larger than $x_{pull}$, respectively. The second rule-based (deterministic) policy $\pi_{msd, 2}$ thus selects one of following actions with probability one:
\begin{align}
\left \{
\begin{array}{r l}
 a_{pull} \; , &\text{if } x_t > x_{pull} \\
 0. \; , &\text{if }  x_{push} \leq x_t \leq x_{pull} \\
 a_{push} \; , &\text{if } x_t < x_{push} \; .
\end{array}
\right . 
\end{align}
After performing some simulations, we empirically observed that the average expected return of the second policy $\pi_{msd, 2}$ over these simulations nearly equals the upper bound discussed previously with the following parameters: $a_{push} = 0$, $x_{pull} = x_{push} = x_{ref}$, and $a_{pull} = -0.3$. In other words, we counterbalance the effect of the spring when the point mass compresses the spring and passes the reference position $x_{ref}$, and otherwise let the system evolve freely:
\begin{align}
\left \{
\begin{array}{r l}
 -0.3 \; , &\text{if } x_t > x_{ref} \\
 0.0 \; , &\text{otherwise } \; .
\end{array}
\right . 
\end{align}

\paragraph{Numerical Values.}  In this work, we will consider the values given in Table \ref{tab:msd_input_param} for the constant parameters.

\begin{table}[H]
	\begin{center}
		\renewcommand\arraystretch{1}
		\caption{Parameters for the MSD.}
		\begin{tabular}[b]{l r}
		\hline
		Symbol & Value \\
		\hline
		$x_{0, min}$ & $0.198$ \\
		$x_{0, max}$ & $0.202$ \\ 
		$v_{0, min}$ & $-0.010$ \\
		$v_{0, max}$ & $0.010$ \\
		$x_{ref}$ & $0.200$ \\
		$c_\omega$ & $0.500$ \\
		$c_\zeta$ & $0.500$ \\
		$K$ & $3.000$ \\
		$c_0$ & $0.500$ \\
		$c_1$ & $-0.300$ \\
		$c_2$ & $0.200$ \\
		$T$ & $100$ \\
		\hline
		\end{tabular}
		\label{tab:msd_input_param}
	\end{center}
\end{table}

\newpage
\section{Optimal Design of a Solar Off-Grid Microgrid} \label{ap:mg}
\begin{figure}[h!]
	\centering
	\includegraphics[width=0.4\linewidth]{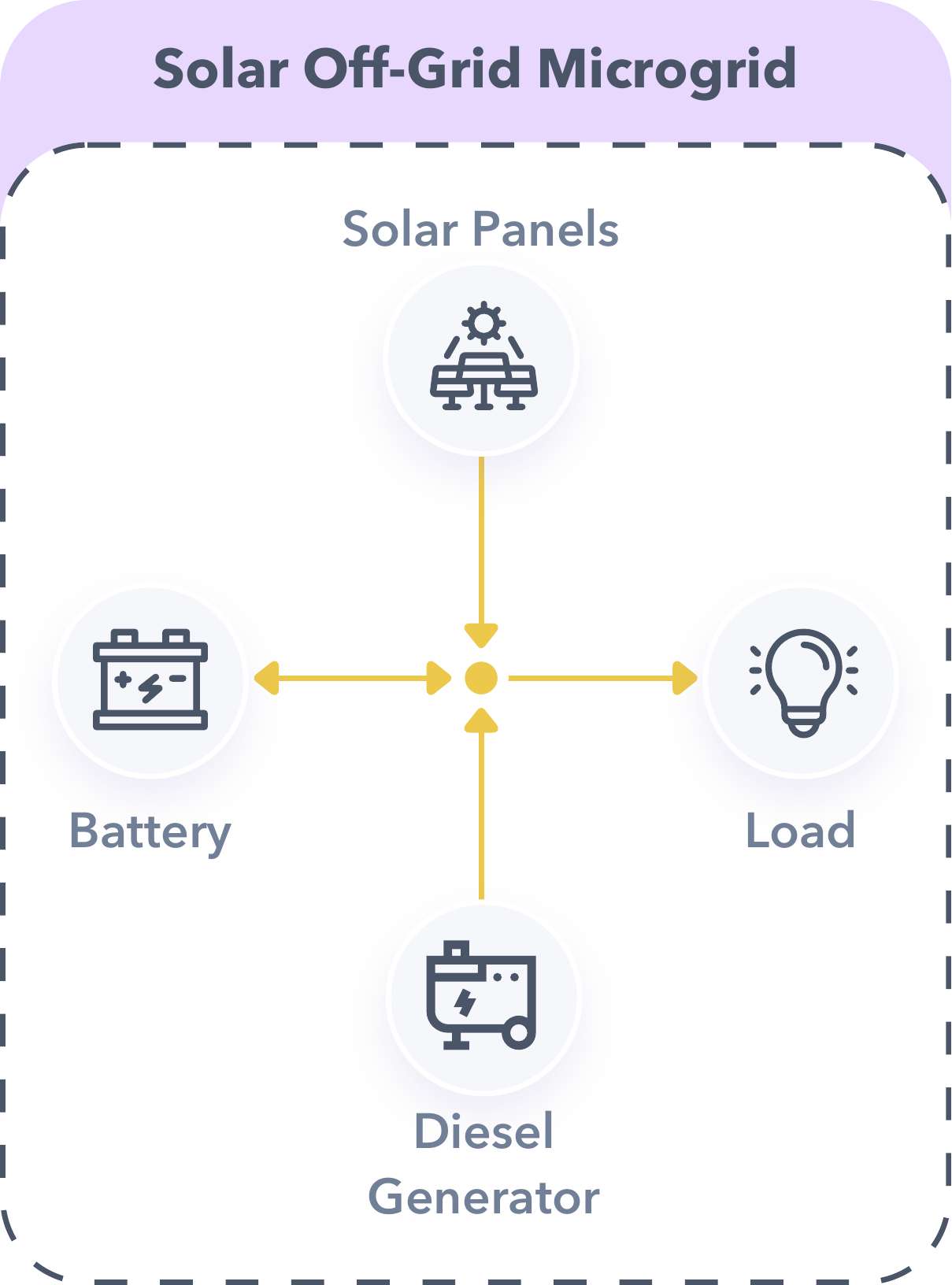}
	\caption{Microgrid configuration.}
	\label{fig: microgrid}
\end{figure}

A solar off-grid microgrid is a small-scale electrical grid composed of photovoltaic (PV) panels converting solar energy into electricity, a battery storing electricity, and a dispatchable diesel generator (genset), which can be switched on and off to supply an electrical load. A sketch of the system configuration considered is presented in Figure \ref{fig: microgrid}. We are interested in optimally sizing and controlling the microgrid, that is, identifying the capacity of components and the control policy that jointly minimize the total cost of the system over its lifetime. The total cost of the microgrid is the sum of the overnight investment costs incurred when installing the various components and the operating costs incurred over the lifetime of the system. The latter include the fuel costs of the genset and the penalties incurred for shedding the load (in the event of a power shortage) and curtailing the electricity generation (if the production surplus cannot be stored). An agent jointly designing and controlling the system faces a trade-off between building a large installation, which typically leads to high investment costs but low (variable) operating costs, and deploying a small installation, which often results in low investment costs but high (variable) operating costs.  

Before formally introducing this benchmark problem, let us mention that we will use the notation $\left[ \cdot \right]$ to indicate the physical unit of the symbol preceding it. In this section, $\left[ W \right]$ denotes instantaneous power production in Watts, $\left[ W_p \right]$ denotes nameplate (manufacturer) power capacity, $\left[ Wh \right]$ denotes energy in Watt-hours and $\left[ Wh_{p} \right]$ denotes nameplate (manufacturer) energy capacity. In addition, $\left[ h \right]$ denotes the unit of time (hours) and $\left[ \$ \right]$ denotes the currency considered in this problem. 

In the following, the solar off-grid microgrid system is modeled as a discrete-time dynamical system that fits into the framework introduced in Section \ref{sec:background}. In addition, the time period $[t, t+1[$ between any two successive time steps $t$ and $t+1$ corresponds to one hour (i.e., $\Delta t = 1$ hour) during which the consumption and the production levels are assumed constant.

\paragraph{Optimization Horizon.} The optimization horizon is denoted as $T=120$, and corresponds to the number of hours over which the system is optimized. The horizon corresponds to a truncation of the system lifetime.

\paragraph{State Space.} The state of the system can be fully described by
$s_t = (SoC_t, h_t, P^{G}_t, \bar{P}^{C}_t, \bar{P}^{PV}_t) \in \St = \left[0,C^B\right] \times \{0,...,23\} \times \R^+ \times \R^+ \times \R^+$, 
where:
\begin{itemize}
	\item $SoC_t \left[Wh \right] \in \left[0, C^B\right]$ denotes the state of charge of the battery at time $t$, with $C^B \left[Wh_p \right] \in \mathbb{R^{+}}$ the installed battery capacity (i.e., the maximum amount of energy that can be stored in the battery).
	\item $h_t \left[h \right] \in \{0,...,23\}$ denotes the hour of the day at time $t$.
	\item $P^{G}_t \left[ W \right] \in \left[0, C^G\right]$ denotes the power produced by the genset at time $t$, where $C^G \left[ W \right] \in \mathbb{R^{+}}$ is the capacity of the generator (i.e., its maximum power output).
	\item $\bar{P}^{C}_t \left[ W \right] \in \mathbb{R^{+}}$ denotes the expected value of the electrical demand at time $t$.
	\item $\bar{P}^{PV}_t \left[ W \right] \in \mathbb{R^{+}}$ denotes the expected value of the PV power generation at time $t$. In addition $C^{PV} \left[W_{p} \right] \in \mathbb{R^{+}}$ denotes the installed capacity of PV panels and $\bar{p}^{PV, h} \left[\% \right]$ is the expected PV production per unit of installed capacity for hour $h$ of the day. The latter is a known parameter given in the last column of Table \ref{tab: mean_cons_pv} for every hour $h$ of the day.
\end{itemize}

\paragraph{Initial State Distribution.} The initial state of charge $SoC_0$ is set to half the capacity of the battery $C^{B}$, the initial hour $h_0$ is set to zero and the generator is initially shut down, such that $P^{G}_0$ is set equal to zero with probability one. The initial value for $\bar{P}^{C}_0$ is given on the first line of Table \ref{tab: mean_cons_pv} in the corresponding column. Finally, the initial value for $\bar{P}^{PV}_0$ is given by the product of $C^{PV}$ and the first entry in column $\bar{p}^{PV, h}$ in Table \ref{tab: mean_cons_pv}.

\paragraph{Action Space.} The available actions correspond to the power exchanged with the battery (in order to charge or discharge it) as well as the power produced by the generator. The former, which is denoted as $\tilde{P}^{B}_t \left[ W \right]$, is assumed to be positive if the battery is being charged and negative if it is being discharged. The latter is denoted as $\tilde{P}^{G}_t \left[ W \right]$. We therefore consider the following continuous action space: 
\begin{align}
\A = \left[-C^{B}, C^{B} \right] \times \left[ 0, C^{G} \right] \; .
\end{align}

\paragraph{Disturbance Space.} The disturbance $\xi_t = E^{C}_t \in \Xi \subseteq \R$ is assumed to represent the stochastic deviation from the expected electricity consumption for hour $h_t$.

\paragraph{Disturbance Distribution.} The disturbance at time $t$ is sampled from a normal distribution centered at zero with standard deviation $\sigma_{C,h}$ depending on the hour of the day $h = h_t$:
\begin{align}
P_\xi(\xi_t|s_t, a_t) &= \mathcal{N}(\xi_t|0, \sigma^{C,h}) \; . 
\end{align}
The values of the standard deviation $\sigma^{C,h}$ are given in Table \ref{tab: mean_cons_pv} for every hour $h$ of the day. 

\paragraph{Transition Function.} The state of charge of the battery $SoC_t$ is updated based on the action $\tilde{P}^{B}_{t}$, which represents the amount of power that we would like to charge into/discharge from the battery during one hour ($\Delta t = 1$h). However, given an action $\tilde{P}^{B}_{t}$, the actual power that can be exchanged with the battery is constrained either by the battery capacity when charging it or by the energy stored in the battery when discharging it. Thus, the actual power exchanged with the battery $P^{B}_{t} \left[W \right] \in \left[-C^{B}, C^{B} \right]$ can be expressed as follows:
\begin{gather}
\begin{split}
P^{B}_{t} =
\begin{cases}
(C^{B}-SoC_{t}) \; ,	& \text{if } \tilde{P}^{B}_{t} > (C^{B}-SoC_{t}) \\
-SoC_{t} \; ,	& \text{if } \tilde{P}^{B}_{t} < -SoC_{t} \\
\tilde{P}^{B}_{t} \; ,	& \text{otherwise}  \; . \label{eqn: clip_battery_power}
\end{cases}
\end{split}
\end{gather}
The state of charge of the battery is then updated using a linear water tank model \citep{boukas2020deep} as follows:
\begin{align}
SoC_{t+1} = SoC_{t}+ \Delta t \cdot P^{B}_{t} \; . \label{eqn: storagedynamics} 
\end{align}
Each time step in the discrete system corresponds to one hour, which is captured by the transition of the state variable $h$:
\begin{gather}
h_{t+1} = (h_t + 1) \mod 24 \; . \label{eqn: time}
\end{gather}
Furthermore, we compute the power produced by the generator at time $t+1$ from the action $\tilde{P}^{G}_t$ as follows:
\begin{align}
P^{G}_{t+1} = \tilde{P}^{G}_t \; . \label{eqn: gendynamics}
\end{align}
The variable $\bar{P}^{C}_{t+1}$ takes the value reported in Table \ref{tab: mean_cons_pv} on the line corresponding to hour $h = h_{t+1}$. Finally, the variable $\bar{P}^{PV}_{t+1}$ is updated as:
\begin{gather}
	\bar{P}^{PV}_{t+1} = \bar{p}^{PV, h_{t+1}} \cdot C^{PV} \; ,
\end{gather} 
where $\bar{p}^{PV, h_{t+1}}$ takes the value reported in Table \ref{tab: mean_cons_pv} in the last column on the line corresponding to hour $h = h_{t+1}$.

\paragraph{Reward Function.} 

The reward signal is, in this case, a cost function composed of two parts, namely the investment cost and the operating cost. The reward signal is given by:
\begin{gather}
r_t = \rho(s_t, a_t, \xi_t) = -(c^{fix}_t + c^{op}_t) \; ,
\end{gather}
where $c^{fix}_t \left[ \$ \right]  \in \mathbb{R^{+}}$ represents a fixed hourly payment that goes towards the settlement of the initial investment cost and $c^{op}_t \left[ \$ \right]  \in \mathbb{R^{+}}$ corresponds to the operating cost at each time step $t$.

In order to compute the fixed-cost term $c^{fix}_t$, we first compute the investment costs of the three components of the microgrid. We assume a small-scale installation where investment costs are quadratic functions of the installed capacity, reflecting diseconomies of scale \citep{friedman2007price}. Let  $c^{PV}_1 \left[ \$/W_{p} \right]  \in \mathbb{R^{+}}$ and $c^{PV}_2 \left[ \$/W_{p}^2 \right]  \in \mathbb{R^{+}}$  denote the cost per unit and per squared-unit of installed PV capacity, respectively. The total installation cost of PV panels $I^{PV} \left[ \$ \right]  \in \mathbb{R^{+}}$ is defined as:
\begin{gather}
I^{PV} =  c^{PV}_1 \cdot C^{PV} +  c^{PV}_2 \cdot {C^{PV}}^2 \; . \label{eqn: pv_inv}
\end{gather}
Let $c^{B}_1 \left[ \$/Wh_{p} \right]  \in \mathbb{R^{+}}$ and $c^{B}_2 \left[ \$/Wh_{p}^2 \right]  \in \mathbb{R^{+}}$ denote the cost per unit and per squared-unit of installed battery capacity, respectively. The total installation cost of the battery storage system $I^{B} \left[ \$ \right]  \in \mathbb{R^{+}}$ is defined as:
\begin{gather}
I^{B} =  c^{B}_1 \cdot C^{B} + c^{B}_2 \cdot {C^{B}}^2 \; . \label{eqn: storage_inv}
\end{gather}
In addition, let $c^{G}_1 \left[ \$/Wh_{p} \right]  \in \mathbb{R^{+}}$ and $c^{G}_2 \left[ \$/Wh_{p}^2 \right]  \in \mathbb{R^{+}}$ denote the cost per unit and per squared-unit of diesel generator capacity, respectively. The total installation cost of the genset $I^{G} \left[ \$ \right]  \in \mathbb{R^{+}}$ is defined as:
\begin{gather}
I^{G} =  c^{G}_1 \cdot C^{G} + c^{G}_2 \cdot {C^{G}}^2 \; . \label{eqn: gen_inv}
\end{gather}
The total investment cost $I \left[ \$ \right]  \in \mathbb{R^{+}}$ is the sum of the investment costs of each component of the microgrid, and is thus defined as:
\begin{gather}
I = I^{B} +I^{PV} + I^{G} \; .
\end{gather}
We assume that this investment is financed via a loan. Hence, an annuity $c^{year} \left[ \$ \right]  \in \mathbb{R^{+}}$ is paid on a yearly basis over the lifetime of the system in order to pay back the principal and interests, which are charged at a yearly rate $r \in \left[0, 1\right]$. This annuity is given by the following accounting formula: 
\begin{gather}
c^{year} = I \frac{r(1+r)^{n}}{(1+r)^{n}-1} \; ,
\end{gather}
where $n \left[\text{year} \right] \in \mathbb{N}$ is the number of years assumed for the economic lifetime of the system. Since a common (non-leap) year has 8760 hours, we define the fixed hourly cost as:
\begin{gather}
c^{fix}_t = \frac{c^{year}}{8760} \; .
\end{gather}

In order to compute the operating cost $c^{op}_t$, we proceed as follows. First, let $P^{C}_t \left[ W \right] \in \mathbb{R^{+}}$ be the realization of the consumption after an action is taken at time $t$, that is:
\begin{gather}
P^{C}_{t} = \bar{P}^{C}_t + E_t^{C} \; . \label{eqn: load}
\end{gather}
Then, we denote by $P^{B, eff}_{t} \left[ W \right]$ the effective power charged or discharged from the battery (i.e., accounting for losses). The latter is computed as follows:
\begin{gather}
\begin{split}
P^{B, eff}_{t} = 
\begin{cases}
 P^{B}_{t} / \eta_{ch} \; ,	& \text{if } P^{B}_{t} \geq 0\\
 \eta_{dis} P^{B}_{t} \; ,	& \text{if } P^{B}_{t} < 0 \; ,
\end{cases}
\end{split}
\end{gather}
where $\eta_{ch}\in \left[0,1\right]$ and $\eta_{dis}\in \left[0,1\right]$ represent the charging and discharging efficiencies of the battery storage system. At each time step $t$ in the horizon, power production and consumption must be balanced. The residual power that may result from any mismatch between production and consumption is denoted as $P^{res}_t \left[ W \right]  \in \mathbb{R}$. Formally, the residual power is given by:
\begin{gather}
P^{res}_t = \bar{P}^{PV}_t + \tilde{P}^{G}_t - P^{C}_{t} - P^{B, eff}_{t} \; . \label{eqn: powerbalance}
\end{gather}
If $P_t^{res}$ is positive, the associated generation surplus is penalized with a cost $\pi^{curtail}\left[ \$/ W \right] \in \R^+$. If $P_t^{res}$ is negative, the associated power shortage is penalized with a cost $\pi^{shed} \left[ \$/ W \right]  \in \mathbb{R^{+}}$. The cost $c^{res}_t \left[ \$ \right] \in \R^+$ resulting from any production-consumption mismatch is therefore computed as:
\begin{gather}
\begin{split}
c^{res}_t = 
\begin{cases}
\pi^{curtail} \cdot P_t^{res} \; , & \text{if } P_t^{res} \geq 0\\
\pi^{shed} \cdot P_t^{res} \; ,	& \text{if } P_t^{res} < 0 \; . \label{eqn: balancecost}
\end{cases}
\end{split}
\end{gather} 
In addition, two types of costs stem from the operation of the diesel generator. First, for each watt produced by the diesel generator, a cost $\pi^{fuel} \left[ \$/ W \right]  \in \mathbb{R^{+}}$ is incurred for buying diesel, such that the fuel cost $c^{fuel}_t \left[ \$ \right]  \in \mathbb{R^{+}}$ at time $t$ is computed as follows:
\begin{gather}
c^{fuel}_t = \pi^{fuel} \cdot \tilde{P}^{G}_t / \eta_{G} \; , \label{eqn: fuelcost}
\end{gather}
where $\eta_{G}\in \left[0,1\right]$ represents the thermal efficiency of the diesel generator. Second, a ramping cost $c^{ramp} \left[ \$ \right]  \in \mathbb{R^{+}}$ is associated with changes in the output of the diesel generator between successive time periods. This cost is assumed to be a quadratic function of the change in power output and takes the following form:
\begin{gather}
\begin{split}
c^{ramp}_t = 
\begin{cases}
\pi^{up} \cdot (\tilde{P}^{G}_t - P^{G}_t)^2 \; , & \text{if } \tilde{P}^{G}_t \geq P^{G}_t\\
\pi^{down} \cdot (\tilde{P}^{G}_t - P^{G}_t)^2 \; ,	& \text{if } \tilde{P}^{G}_t < P^{G}_t \; , \label{eqn: rampcost}
\end{cases}
\end{split}
\end{gather} 
where $\pi^{up} \left[ \$/ W^2 \right]  \in \mathbb{R^{+}}$ and $\pi^{down} \left[ \$/ W^2 \right]  \in \mathbb{R^{+}}$ represent the cost of increasing and decreasing the power output of the generator from time $t$ to time $t+1$, respectively. Finally, the total operating cost is the sum of the production-consumption mismatch, fuel and ramping costs:
\begin{gather}
c^{op}_t = c^{res}_t + c^{fuel}_t + c^{ramp}_t \; . \label{eqn: opcost}
\end{gather}

\paragraph{Parametrized Environment.} The off-grid microgrid environment is the 8-tuple \\ $(\St, \A, \Xi, P_0, f_\psi, \rho_\psi, P_\xi, T)$, which is parametrized by the vector $\psi = (C^{B}, C^{PV}, C^{G}) \in \mathbb{R^{+}}^{3}$. We note that, strictly speaking, the transition and reward functions of this environment are not differentiable. First, the state variable $h_t$ takes values in a discrete set and its update involves modular arithmetic. However, this update is independent of the environment parameters. In this case, the state variable $h_t$ is thus considered as a constant by the gradient operator and does not participate in the update of the environment parameters when doing gradient ascent/descent. Second, the clipping operations (e.g., equation \eqref{eqn: clip_battery_power}) that depend on state variables and on the environment parameters are also locally non-differentiable. In the latter case, gradients are replaced by subgradients when updating parameters. In our implementation, PyTorch handles this automatically.

\paragraph{Rule-Based Policies.} We propose two rule-based control policies for the microgrid system. The first one seeks to greedily minimize the average operating cost. More precisely, the solar panels are first used to supply as much of the expected demand as possible at zero marginal cost. If the power production from PV panels exceeds the expected demand, the battery is charged with the production surplus and the diesel generator remains unused. On the other hand, if the expected demand exceeds the PV production, the battery is discharged to compensate for the power shortage. If the combined output from the PV panels and the battery is insufficient to supply the demand in full, the diesel generator is activated. Formally, given the state $s_t$, the first (deterministic) rule-based policy $\pi_{mg, 1}$ selects one of the following actions with probability one:
\begin{equation}
\left \{
\begin{array}{r l}
 (\min(C^{B} - SoC_t, \bar{P}^{PV}_{t} - \bar{P}^{C, h_t}), 0 ) \; , &\text{if } \bar{P}^{C, h_t} \leq \bar{P}^{PV}_{t} \\
 (-SoC_t , \min(C^{G},  \bar{P}^{C, h_t} - \bar{P}^{PV}_{t} - SoC_t ) \; , &\text{if } \bar{P}^{C, h_t} \geq \bar{P}^{PV, h_t} + SoC_t \\
(\max(-SoC_t, \bar{P}^{PV}_{t} - \bar{P}^{C, h_t}), 0 ) \; , &\text{otherwise} \; .
\end{array}
\right . 
\end{equation}
This policy neglects the ramping costs of the diesel generator and may lead to large operational costs for systems with small batteries and large diesel generators. An alternative consists in using the diesel generator at full capacity in order to provide a constant power supply and using the PV panels as well as the battery to balance the production and the consumption. Formally, given the state $s_t$, the second (deterministic) rule-based policy $\pi_{mg, 2}$ selects one of the following actions with probability one:
\begin{align}
\left \{
\begin{array}{r l}
(\min(C^{B} - SoC_t, \bar{P}^{PV}_{t} + C^{G} - \bar{P}^{C, h_t}), C^{G}) \; , &\text{if } \bar{P}^{C, h_t} - C^{G} \leq \bar{P}^{PV}_{t} \\
(\max(-SoC_t, \bar{P}^{PV}_{t} + C^{G} - \bar{P}^{C, h_t}), C^{G} ) \; , &\text{if } \bar{P}^{C, h_t} - C^{G} \geq \bar{P}^{PV}_{t} \; .
\end{array}
\right . 
\end{align}

\paragraph{Numerical values.} Table \ref{tab: input_param} summarizes the parameter values used in the microgrid experiments presented in this paper.
\begin{table}[H]
	\begin{center}
		\renewcommand\arraystretch{1}
		\caption{Parameters used for the solar off-grid microgrid.}
		\begin{tabular}[b]{l c c}
			\hline
			Symbol & Value & Unit\\
			\hline
			$\eta_{ch}$ 	& 1 & - \\
			$\eta_{dis}$ 	& 1 & - \\
			$\eta_{G}$ 		& 1 & - \\
			$c^{PV}_1$		& 200 & $\$/W_{p}$ \\
			$c^{PV}_2$		& 100 & $\$/W_{p}$ \\
			$c^{B}_1$		& 100 & $\$/W_{p}$ \\
			$c^{B}_2$		& 20 	& $\$/W_{p}$ \\
			$c^{G}_1$		& 1000 	& $\$/W_{p}$ \\
			$c^{G}_2$		& 10000 & $\$/W_{p}$ \\
			$r$				& 0.1 & - \\
			$n$				& 20 	& years\\
			$\pi^{shed}$ 	& 25 	& $\$/Wh$ \\
			$\pi^{curtail}$	& 25 	& $\$/Wh$ \\
			$\pi^{fuel}$ 	& 4 	& $\$/Wh$ \\
			$\pi^{up}$ 		& 0.5 	& $\$/Wh$ \\
			$\pi^{down}$	& 0 	& $\$/Wh$ \\
			$T$ 			& 120 	& hour \\
			$\Delta t$ 		& 1 	& hour \\
			\hline
		\end{tabular}
		\label{tab: input_param}
	\end{center}
\end{table}
\begin{table}[H]
	\begin{center}
		\renewcommand\arraystretch{1}
		\caption{Electrical load consumption and PV production capacity factor data.}
		\begin{tabular}[b]{l c c c}
			\hline
			Hour $h$ & $\bar{P}^{C, h}$ & $\sigma^{C,h}$ & $\bar{p}^{PV, h}$\\
			\hline
			0& 10.4 &0.55& 0.0\\
			1& 9.7 &0.50& 0.0\\
			2& 9.3 &0.43& 0.0\\
			3& 8.9 &0.39& 0.0\\
			4& 8.6 &0.39& 0.0\\
			5& 8.2 &0.37& 0.0\\
			6& 7.3 &0.37& 0.0\\
			7& 6.8 &0.36& 0.0\\
			8& 6.9 &0.40& 0.0\\
			9& 7.0 &0.43& 0.04\\
			10& 7.2 &0.44& 0.08\\
			11& 7.4 &0.47& 0.12\\
			12& 7.7 &0.42& 0.14\\
			13& 8.0 &0.40& 0.15\\
			14& 8.2 &0.42& 0.14\\
			15& 8.2 &0.47& 0.12\\
			16& 8.1 &0.43& 0.08\\
			17& 8.8 &0.44& 0.04\\
			18& 12.6 &0.81& 0.0\\
			19& 16.0 &0.60& 0.0\\
			20& 16.5 &0.55& 0.0\\
			21& 15.8 &0.57& 0.0\\
			22& 13.9 &0.60& 0.0\\
			23& 11.8 &0.59& 0.0\\
			\hline
		\end{tabular}
		\label{tab: mean_cons_pv}
	\end{center}
\end{table}

\newpage
\section{Drone Design and Control} \label{ap:drone}
\begin{figure}[H]
    \centering
    \includegraphics[width=0.8\linewidth]{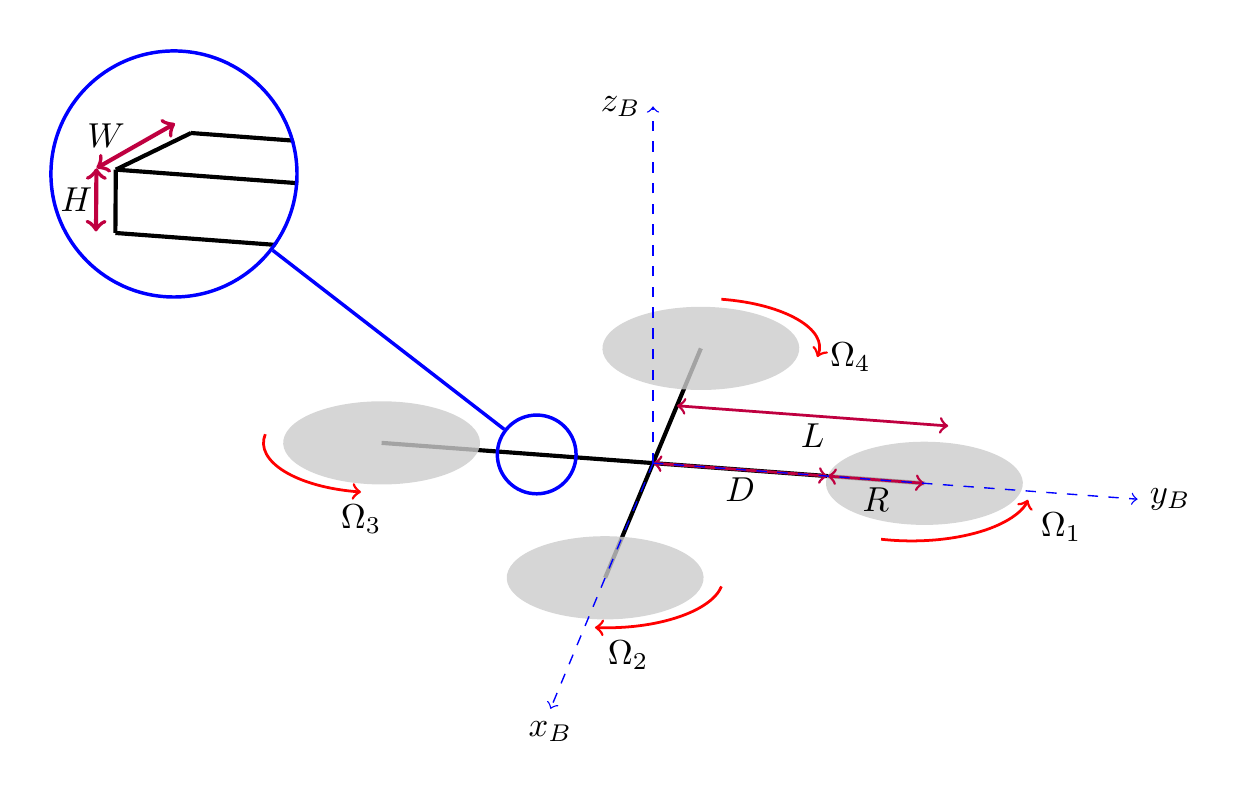}
    \caption{Schematic representation of a quadrotor.}
    \label{fig:drone}
\end{figure}

This section formulates the problem of jointly designing and controlling a drone in the framework of Section \ref{sec:background}. More specifically, the drone design considered here is a so-called quadrotor, which relies on four independently actuated propellers to fly, as shown in Figure~\ref{fig:drone}. Such systems have been studied extensively in recent years \citep{habib2014dynamic, sabatino2015quadrotor, hwangbo2017control, wang2017autonomous, hodge2021deep}, but mostly from a control perspective.

The quadrotor is modelled as a rigid body moving in three-dimensional space. The quadrotor possesses six degrees of freedom, namely three translations and three rotations. Hence, the position of the quadrotor is described by six variables, which are expressed using two different references frames. The first reference frame, which is called the \textit{earth frame}, is an inertial frame used to measure the position of the center of mass of the quadrotor. In this paper, the center of mass and the barycenter of the drone are assumed to coincide. The second reference frame, which is referred to as the \textit{body frame}, is attached to the barycenter of the quadrotor and its axes are parallel to the quadrotor's principal axes of inertia. This frame is used to measure the rotation of the quadrotor around its barycenter. The quadrotor is actuated via the angular speed of its four propellers, and the system is therefore underactuated. The equations of motion are derived using Newton's second law and Euler's rotation equations, resulting in a system of highly nonlinear second-order ordinary differential equations that describe the dynamics of the system in time. The design of the quadrotor is defined by four physical parameters, namely the length, the thickness and the width of the arms connecting its propellers to its barycenter, as well as the radius of its propellers. The goal is then to design a quadrotor that can move as fast as possible around an elliptical trajectory while being subject to random wind gusts. This objective involves a trade-off between the inertia of the drone and its controllability. Indeed, a heavy drone will typically be much less sensitive to wind gusts than a light one is but increasing its weight and size will also make it more difficult to control and maneuver (and vice-versa).

\paragraph{Optimization Horizon.} The trajectory of the drone is optimized over $T=100$ time steps. The time elapsed between two successive time steps $t$ and $t+1$ is denoted as $\Delta t$ and reported in Table \ref{tab:drone-input-param}, resulting in a total flight time of seven seconds.

\paragraph{State Space.} The state space model has twelve state variables. These variables represent the three angular positions and the three linear positions in the earth frame $(\phi_t, \theta_t, \varphi_t, x_t, y_t, z_t)$, as well as the three angular velocities and the three linear velocities in the body frame $(p_t, q_t, r_t, u_t, v_t, w_t)$ at time $t$ (i.e., $t \times \Delta t$ seconds into the flight):
\begin{align}
    s_t = (\phi_t, \theta_t, \varphi_t, p_t, q_t, r_t, u_t, v_t, w_t, x_t, y_t, z_t) \in \mathcal{S} = \mathbb{R}^{12} \; .
\end{align}
 
\paragraph{Initial State Distribution.} The drone initially starts at rest at the origin of the earth frame with probability one:
\begin{align}
    s_0 &= (0, 0, 0, 0, 0, 0, 0, 0, 0, 0, 0, 0) \; .
\end{align}

\paragraph{Action Space} The drone is controlled through the angular speeds of its propellers:
\begin{align}
    a_t = (\Omega_1, \Omega_2, \Omega_3, \Omega_4) \in \mathcal{A} = \left [\Omega_{\min}, \Omega_{\max} \right ]^4 \subsetneq \mathbb{R}^{4} \; .
\end{align}

\paragraph{Disturbance Space.} The disturbance space consists of the three forces (in the body frame) and three torques (also in the body frame) applied on the drone by the wind, such that:
\begin{align}
    \xi_t = (f_{wx}, f_{wy}, f_{wz}, \tau_{wx}, \tau_{wy}, \tau_{wz}) \in \Xi = \mathbb{R}^6 \; .
\end{align}

\paragraph{Disturbance Distribution.} Let us assume that the force applied by the wind on the center of mass of the drone follows a multivariate normal distribution with mean $\mu_w \mathbf{1} $ and covariance matrix $\sigma_w I$, with $\mathbf{1} \in \mathbb{R}^3$ a vector with all entries equal to $1$ and $I \in \mathbb{R}^{3\times 3}$ the identity matrix of size three. In addition, let us assume that the wind applies no torque on the drone. Hence, we get:
\begin{align}
    (f_{wx}, f_{wy}, f_{wz}) &\sim \mathcal{N}(\cdot| \mu_w\mathbf{1}, \sigma_w^2 I) \\
    (\tau_{wx}, \tau_{wy}, \tau_{wz}) &= 0 \;.
\end{align}
\paragraph{Transition Function.} The model used in this section is adapted from the work of \cite{sabatino2015quadrotor}, and we only briefly review its derivation. The equations of motion of the drone are derived from Newton's second law and Euler's rotation equations, which can be expressed as vectorial second-order ordinary differential equations where the dependent variables are the linear and angular positions of the drone, respectively. These differential equations represent how the linear and angular positions evolve over time in response to the application of forces and torques on the drone (by its own propellers and wind gusts), and can be equivalently expressed as a system of first-order differential equations with twelve scalar variables. These two laws, which are stated in the body frame, only provide six equations. Thus, six equations describing the kinematics of the drone (i.e., linking the linear and angular velocities in the earth and body frames) are added in order to obtain the required twelve equations. These equations are then discretized in time order to obtain the state space model. We start with the equations modeling the dynamics of the drone and then proceed with the description of its kinematics.

Let $f_{tr}$ be the total thrust generated by the propellers, let $(f_{wx}, f_{wy}, f_{wz})$ be the forces produced by wind on the quadrator, let $(\tau_x, \tau_y, \tau_z)$ denote the control torques generated by the propellers and let $(\tau_{wx}, \tau_{wy}, \tau_{wz})$ be the torques produced by the wind. In addition, let $g$ and $m$ be the gravitational acceleration and the mass of the drone, and let $(I_x, I_y, I_z)$ be the moments of inertia around the principal axes of the drone. Then, applying Euler's rotation equations and Newton's second law in the body frame yields the following equations:
\begin{align}
    \dot p &= \frac{I_y - I_z}{I_x} r q + \frac{\tau_x +\tau_{wx}}{I_x} \\
    \dot q &= \frac{I_z - I_x}{I_y} p r + \frac{\tau_y +\tau_{wy}}{I_y} \\
    \dot r &= \frac{I_x - I_y}{I_z} p q + \frac{\tau_z +\tau_{wz}}{I_z} \\
    \dot u &= r v - q w - g \sin \theta + \frac{f_{wx}}{m} \\
    \dot v &= p w - r u + g \sin \phi \cos \theta + \frac{f_{wy}}{m} \\
    \dot w &= q u - p v + g \cos \theta \cos \phi + \frac{f_{wz} - f_{tr}}{m} \; .
\end{align}
Note that the force $f_{tr}$ and the torques $(\tau_x, \tau_y, \tau_z)$ produced by the propellers can be expressed in terms of their angular speeds $(\Omega_1, \Omega_2, \Omega_3, \Omega_4)$. More precisely, let $b$ and $d$ be the thrust and drag factors of the drone, and let $L$ be the length of its arms. Then, the force and torques can be expressed as:
\begin{align}
    \begin{bmatrix}
    f_{tr}\\
    \tau_x\\
    \tau_y\\
    \tau_z
    \end{bmatrix}
    &=
    \begin{bmatrix}
    b & b & b & b \\
    -bL & 0 & b L & 0\\
    0 & -b L &0 & b L\\
    -d & d & -d & d
    \end{bmatrix}
    \begin{bmatrix}
    \Omega_1^2\\
    \Omega_2^2\\
    \Omega_3^2\\
    \Omega_4^2
    \end{bmatrix} \; .
\end{align}

The kinematic model of the drone relies on a rotation matrix to link the linear velocities in the earth frame to the ones in the body frame:
\begin{align}
    \begin{bmatrix}
    \dot x\\
    \dot y \\
    \dot z
    \end{bmatrix}
    &=
    \begin{bmatrix}
    \cos \theta \cos \varphi & \sin \phi \sin \theta \cos \varphi - \cos \phi \sin \varphi & \cos \phi \sin \theta \cos \varphi + \sin \phi \sin \varphi\\
    \cos \theta \sin \varphi & \sin \phi \sin \theta \sin \varphi + \cos \phi \cos \varphi & \cos \phi \sin \theta \sin \varphi - \sin \phi \cos \varphi\\
    - \sin \varphi & \sin \phi \cos \theta & \cos \phi \cos \theta
    \end{bmatrix}
    \begin{bmatrix}
    u\\
    v \\
    w
    \end{bmatrix} \; ,
\end{align}
while an angular transformation matrix is used to link the angular velocities in the earth frame to the ones in the body frame:
\begin{align}
    \begin{bmatrix}
    \dot \phi\\
    \dot \theta \\
    \dot \varphi
    \end{bmatrix}
    &=
    \begin{bmatrix}
    1 & \sin \phi \tan \theta & \cos \phi \tan \theta \\
    0 & \cos \phi & -\sin \phi\\
    0 & \frac{\sin \phi}{\cos \theta} & \frac{\cos \phi}{\cos \theta}
    \end{bmatrix}
    \begin{bmatrix}
    p\\
    q\\
    r
    \end{bmatrix} \; .
\end{align}

Finally, these ordinary differential equations are discretized using the forward Euler method to yield a discrete-time state space model. In order to simulate the drone dynamics as accurately as possible, $n_e$ Euler integration steps are taken between successive time periods $t$ and $t+1$. Note that both the control input and the disturbance remain constant during the Euler integration between successive time periods and their values are set to $a_t$ and $\xi_t$, respectively.

\paragraph{Reward Function.} In this problem, the objective is to move the drone around an elliptical trajectory as quickly as possible. In other words, we seek to minimize the distance to the ellipse while maximizing the speed of the drone in the direction tangent to the ellipse.

More formally, let $\mathcal{E}$ be set the of points describing the desired trajectory, and let $r_x \in \R^+$ and $r_y \in \R^+$ denote the lengths of the two axes of symmetry of the associated ellipse. Then, the ellipse $\mathcal{E}$ is given by the points $(x, y, z) \in \mathbb{R}^3$ satisfying the following equations:
\begin{align}
	\mathcal{E} &: \left\{
      \begin{array}{ccl}
        \frac{(x - r_x)^2}{r_x^2} +\frac{y^2}{r_y^2} &=& 1 \\
        z &=& 0 \; . \\
      \end{array}
    \right. 
\end{align}
Let $p_{d, t} = (x_t, y_t, z_t) \in \mathbb{R}^3$ be the linear position of the drone in the earth frame at time $t$ and let $p_{e,t}  = (x_{e,t}, y_{e,t}, z_{e,t}) \in \mathbb{R}^3$ be the projection of $p_{d, t}$ on the ellipse $\mathcal{E}$:
\begin{align}
	p_{e,t} \in \text{argmin}_{x \in \mathcal{E}} ||x - p_{d, t}|| \; ,
\end{align}
where $||\cdot||$ denotes the Euclidean norm. The coordinates of $p_{e, t}$ in the earth frame can be computed in terms of the coordinates of $p_{d, t}$ as follows:
\begin{align}
	\left\{
      \begin{array}{ccl}
        x_{e,t} &=& \text{sign}(y_t) (x_t - r_x) \frac{r_x}{\sqrt{r_x ^2 + (x_t - r_x)^2}}  + r_x \\
        y_{e,t} &=& \text{sign}(y_t) \frac{r_x}{\sqrt{\big(\frac{r_x}{r_y}\big)^2 + \big(\frac{x_t - r_x}{y_t}\big)^2}} \\
        z_{e,t} &=& 0 \; , \\
      \end{array}
    \right. 
\end{align}
which holds true if $y_t$ is non-zero. If $y_t$ is equal to zero, on the other hand, the projection of $p_{d, t}$ on the ellipse is either $(0, 0, 0)$ or $(2r_x, 0, 0)$, depending on the sign of $x_t$. 

In addition, let $\gamma(p_{e, t}) = (\gamma_{x, t}, \gamma_{y, t}, \gamma_{z, t})$ be the tangent vector of $\mathcal{E}$ at $p_{e, t}$, which has the following coordinates in the earth frame:
\begin{align}
	\left\{
      \begin{array}{ccl}
        \gamma_{x, t} &=& r_x \frac{y_{e,t}}{r_y} \\
        \gamma_{y, t} &=& -r_y \frac{x_{e, t} - r_x}{r_x} \\
        \gamma_{z, t} &=& 0 \; . \\
      \end{array}
    \right. 
\end{align}
Then, the reward function can be expressed as:
\begin{equation}
r_t =    - ||p_{d, t} - p_{e,t}||^2 + \lambda \max \Big(v_{\max}, p_{d, t} \cdot \frac{ \gamma(p_{e, t})}{|| \gamma(p_{e, t})||}\Big) \; ,
\end{equation}
where the first term measures the distance from the ellipse and the second term quantifies the speed at which the drone travels around the ellipse. The scalar parameter $\lambda \in \mathbb{R}^+$ controls the extent to which one objective is preferred over the other. In addition, in the second term, the speed of the drone in the direction given by $\gamma(p_{e, t})$ is clipped if it exceeds a threshold speed $v_{\max}$. This provides better control over the relative contributions of both terms in the objective.

Using the coordinates of the various vectors, the reward function can be equivalently written as:
\begin{multline}
    r_t = - \Big( (x_{e, t} - x_t)^2 + (y_{e, t} - y_t)^2 + z_t^2 \Big) \\
     + \lambda \max \big( v_{\max}, \frac{r_x r_y}{\sqrt{r_x^4 y_{e, t}^2 + r_y^4 (x_{e, t} - r_x)^2}} \big ( \dot x_t r_x \frac{y_{e, t}}{r_y} - \dot y_t r_y \frac{x_{e, t} - r_x}{r_x} \big ) \; .
\end{multline}

\paragraph{Parametrized Drone Environment.} Let $D$ be the distance from the barycenter of the drone to the tip of the propellers (see Figure \ref{fig:drone}). Let $H$ and $W$ be the thickness and the width of the arms connecting the barycenter of the drone to the center of its propellers, respectively. In addition, let $R$ be the radius of the propellers. Then, the drone environment is the 8-tuple $(\St, \A, \Xi, P_0, f_\psi, \rho_\psi, P_\xi, T)$, which is parametrized by the vector $\psi = (D, R, H, W) \in {\R^{+}}^4$. The distance from the barycenter of the drone to the center of the propellers $L$, the mass $m$, the moments of inertia $(I_x, I_y, I_z)$, as well as the thrust factor $b$ and the drag factor $d$ depend directly on these parameters. We derive their expressions next. 

First, as can be seen in Figure \ref{fig:drone}, the distance from the barycenter of the drone to the center of the propellers $L$ can be expressed as the sum of the distance from the barycenter to the tip of the propellers $D$ and the radius of the propellers $R$:
\begin{align}
	L = D + R \; .
\end{align}
In the following, for the sake of clarity, expressions will be directly written in terms of $L$ where appropriate. Note that these expressions can be written solely in terms of $D$ and $R$ by substitution.

Second, the mass of the drone can be computed as the integral of the density of the material used to construct the drone $\rho$ over its volume $\Sigma$. The latter essentially corresponds to the volume of the four arms and the joint linking them. By symmetry, it can be computed as the sum of two terms. First, we count the mass of an arm connecting the barycenter of the drone to a propeller four times. Then, we add the mass of the joint and get the total mass of the drone as follows:
\begin{alignat}{2}
    m = \int_\Sigma \rho\: d \sigma 
    &= 4 \int_{-W / 2}^{W / 2} \int_{W / 2}^{L} \int_{-H / 2}^{H / 2} \rho\: dx dy dz 
    &&+ \int_{-W / 2}^{W / 2} \int_{-H / 2}^{H / 2} \int_{-H / 2}^{H / 2} \rho\: dx dy dz \\
    &= 4 \rho H W (L - \frac{W}{2}) 
    && + \rho H W^2 \; .
\end{alignat}
Note that the mass of the propellers is neglected.

Third, by definition, $I_x$ can be computed as the integral of the product of the density with the square of the distance $\sigma$ from the center of gravity to each point of the structure projected on the $yz$-plane $\Sigma_{yz}$:
\begin{align}
    I_x &= \int_{\Sigma_{yz}} \rho \sigma^2\: d \sigma =  \int_{-L}^{L} \int_{-H/2}^{H/2} \rho (y^2+z^2)\: dy dz = \frac{1}{6} \rho L H (4L^2 + H^2) \; .
\end{align}
By symmetry, the moments of inertia $I_x$ and $I_y$ around the $x$ and $y$ principal axes are equal, thus:
\begin{align}
    I_y = I_x = \frac{1}{6} \rho L H (4L^2 + H^2) \; .
\end{align}
By definition, the moment of inertia $I_z$ is computed as follows:
\begin{align}
    I_z &= \int_{\Sigma_{xy}} \rho \sigma^2\: d \sigma\; .
\end{align}
Let us decompose the projected surface $\Sigma_{xy}$ into the four arms and the joint linking them. By symmetry, the moments of inertia of the four arms are equal. We thus have that:
\begin{alignat}{2}
    I_z &= 4 \int_{-W/2}^{W/2} \int_{W/2}^{L} \rho (x^2+y^2)\: dx dy &&+ \int_{-W/2}^{W/2} \int_{-W/2}^{W/2} \rho (x^2+y^2)\: dx dy \\
    &=  \frac{1}{3} \rho W (4L^3 + LW^2 - W^3) &&+ \frac{1}{6} \rho W^4 \; .
\end{alignat}

Finally, we can compute the thrust factor $b$ and the drag factor $d$ as follows \citep{habib2014dynamic}:
\begin{align}
    b &= \frac{1}{2} \rho_{air} C_b A R^2 \\
    d &= \frac{1}{2} \rho_{air} C_d A R^2 \\
    A &= \pi R^2 \; ,
\end{align}
where $C_b$ and $C_d$ are two aerodynamical parameters, $\rho_{air}$ is the density of air and $A$ is the propeller disk area.

\paragraph{Numerical Values.} Table \ref{tab:drone-input-param} summarizes the parameter values used in the drone experiments presented in this paper.
\begin{table}[H]
	\begin{center}
		\renewcommand\arraystretch{1}
		\caption{Parameters used for the drone.}
		\begin{tabular}[b]{l c c}
			\hline
			Symbol & Value & Unit\\
			\hline
			$\rho_{air}$ & $1.225$ & $kg/m^3$ \\
			$\rho$ (Polypropylene) & $900$ & $kg/m^3$ \\
			$C_b$ & $1.0$ & $-$ \\
			$C_d$ & $1.0$ & $-$ \\
			$\lambda$ & $0.3$ & $m^2/(m/s)$ \\
			$\mu_w$ & $0.0$ & $N$ \\
			$\sigma_w$ & $0.003$ & $N$ \\
			$n_e$ & $1$ & $-$ \\
			$\Delta t$ & $0.07$ & $s$ \\
			$v_{\max}$ & $1.0$ & $m/s$ \\
			$r_x$ & $1.0$ & $m$ \\
			$r_y$ & $1.5$ & $m$ \\
			$\Omega_{\min}$ & $0.0$ & $rad/s$ \\
			$\Omega_{\max}$ & $20.0$ & $rad/s$ \\
			\hline
		\end{tabular}
		\label{tab:drone-input-param}
	\end{center}
\end{table}

\newpage
\section{Optimization Parameters of the JODC Algorithm} \label{ap:optim_parameters}
In this section, we provide the parameters used in each experiment involving the JODC algorithm \citep{schaff2019jointly}. For each environment, we used the same policy hypothesis space as the one used for the DEPS algorithm. Similarly, the same scaling and clipping were used. The parametric distribution over environments was chosen to be a normal distribution, parametrized by its mean and its standard deviation. Additional parameters specific to the JODC algorithm are provided in Table \ref{tab:msd-jodc-param}, \ref{tab:mg-jodc-param} and \ref{tab:drone-jodc-param} for the MSD, the microgrid and the drone environments, respectively.

\begin{table}[H]
	\begin{center}
		\renewcommand\arraystretch{1}
		\caption{JODC parameters for the MSD environment.}
		\begin{tabular}[b]{l c c}
			\hline
			Parameter & Experiment 1 & Experiment 2\\
			\hline
			Step size policy parameters & $0.001$ & $0.001$ \\
			Step size environment parameters & $0.005$ & $0.005$ \\
			Batch size $M$ & $64$ & $4$ \\
			PPO epochs & $5$ & $5$ \\
			PPO epsilon clip & $0.1$ & $0.1$ \\
			\hline
		\end{tabular}
		\label{tab:msd-jodc-param}
	\end{center}
\end{table}

\begin{table}[H]
	\begin{center}
		\renewcommand\arraystretch{1}
		\caption{JODC parameters for the microgrid environment.}
		\begin{tabular}[b]{l c}
			\hline
			Parameter & Value\\
			\hline
			Step size policy parameters & $0.001$ \\
			Step size environment parameters & $0.001$ \\
			Batch size $M$ & $64$ \\
			PPO epochs & $5$ \\
			PPO epsilon clip & $0.1$ \\
			\hline
		\end{tabular}
		\label{tab:mg-jodc-param}
	\end{center}
\end{table}

\begin{table}[H]
	\begin{center}
		\renewcommand\arraystretch{1}
		\caption{JODC parameters for the drone environment.}
		\begin{tabular}[b]{l c}
			\hline
			Parameter & Value\\
			\hline
			Step size policy parameters & $0.00005$ \\
			Step size environment parameters & $0.0005$ \\
			Batch size $M$ & $64$ \\
			PPO epochs & $5$ \\
			PPO epsilon clip & $0.1$ \\
			\hline
		\end{tabular}
		\label{tab:drone-jodc-param}
	\end{center}
\end{table}

\newpage
\bibliography{source}

\end{document}